\documentclass[10pt,twocolumn,letterpaper]{article}

\usepackage[pagenumbers]{cvpr}

\usepackage{wrapfig}
\usepackage{tabularx}
\usepackage{float}
\usepackage{needspace}
\usepackage{placeins}
\usepackage{stfloats}
\DeclareCaptionLabelFormat{continued}{#1~#2~(continued)}
\newcommand{\suppdelta}[2]{#1\kern0.08em\textsubscript{\fontsize{5.5}{6}\selectfont\color{black!65}#2}}

\newcommand{\TaskPanel}[3][0.75\linewidth]{%
    \begin{subfigure}{#1}
        \centering
        \IfFileExists{#2}{%
            \includegraphics[clip,width=\linewidth]{#2}%
        }{%
            \fbox{\parbox[c][0.12\textheight][c]{0.92\linewidth}{%
                \centering\small
                Missing task figure. Upload the existing\par
                \texttt{fig/task24} folder before compiling.
            }}%
            \PackageWarningNoLine{MSIU-Supplement}{Missing task figure: #2}%
        }%
        \caption{}%
        \label{#3}%
    \end{subfigure}%
}
\newcommand{\TaskPanelGap}{\par\addvspace{2pt}}
\newenvironment{TaskFigure}[1][p]{%
    \begin{figure*}[#1]
    \centering
    \captionsetup{skip=4pt}
    \captionsetup[subfigure]{skip=2pt}
}{%
    \end{figure*}
}

\newcommand{\ResultParagraph}[1]{%
    \par\addvspace{3pt}%
    \noindent\textbf{#1}\quad\ignorespaces
}

\newcommand{\WideFigurePage}[1]{%
    \clearpage
    
    \twocolumn[{%
      \begin{minipage}{\textwidth}
        \centering
        \captionsetup{type=figure,skip=5pt}
        #1
      \end{minipage}\vspace{12pt}%
    }]
}

\definecolor{cvprblue}{rgb}{0.21,0.49,0.74}
\usepackage[pagebackref,breaklinks,colorlinks,allcolors=cvprblue]{hyperref}
\hypersetup{
  pdftitle={MSIU-Bench: Evaluating and Improving Multimodal Social Interaction Understanding in Multimodal LLMs},
  pdfauthor={Feiyu Du, Xi He, Xinpeng Li, Shijian Deng, Yapeng Tian, Weili Wu}
}

\def\confName{CVPR}
\def\confYear{2027}

\title{MSIU-Bench: Evaluating and Improving Multimodal Social Interaction Understanding in Multimodal LLMs}
\author{%
  Feiyu Du$^{1}$ \quad Xi He$^{1}$ \quad Xinpeng Li$^{1}$ \quad
  Shijian Deng$^{1}$ \quad
  Yapeng Tian$^{1}$\thanks{Corresponding author.} \quad
  Weili Wu$^{1}$\\[0.3em]
  $^{1}$University of Texas at Dallas
}
\date{}

\usepackage{booktabs,tabularx,multirow,array}
\usepackage{xcolor,colortbl}
\usepackage{pifont}

\usepackage{amssymb}
\providecommand{\msiuyes}{%
  \textcolor{green!45!black}{\ensuremath{\checkmark}}}
\providecommand{\msiuno}{%
  \textcolor{red!75!black}{\ensuremath{\times}}}

\begin{document}

\maketitle

\begin{abstract}

Understanding a conversation in a video requires tracking who is speaking to whom, who is being referred to, and how participants' roles change as the conversation unfolds. We introduce \textbf{MSIU-Bench} (\textbf{M}ultimodal \textbf{S}ocial \textbf{I}nteraction
\textbf{U}nderstanding Benchmark), a dataset and benchmark for evaluating and improving these abilities in multimodal large language models (MLLMs). We group 24 tasks into three capability levels according to their focus and temporal scope: social signal
perception and identity grounding, current social interaction and
referential reasoning, and dynamic interaction-state reasoning. MSIU-Bench contains 40,417 question--answer pairs
across 4,260 video clips, with 28,332 pairs in the
training split. We evaluate 21 multimodal models on 8,116 test questions through three complementary
components: Core, Binary, and description-based
multiple-choice questions (Description-MCQ). Gemini-4-Argon, a frontier MLLM, leads the direct-inference baselines with a Core task-macro accuracy of 66.35\%. Fine-grained analysis shows that it still makes errors in speaker--utterance grounding and often misses shifts in addressee from the camera wearer to another participant. After supervised LoRA adaptation on the MSIU-Bench
training set, four open-weight models achieve absolute gains of 25.36\%--30.81\% in Core task-macro accuracy. In particular, D-ORCA-8B-0210 improves from 37.59\% to 62.95\%, bringing it within 3.40\% of Gemini-4-Argon on this metric. MSIU-Bench provides
a data and evaluation foundation for assessing and improving
multimodal social interaction understanding, supporting research
toward more socially capable intelligent systems. Our dataset, benchmark, and pre-trained models will be publicly available.

\end{abstract}

\begin{figure*}[t]
    \centering
    \includegraphics[width=0.9\textwidth]{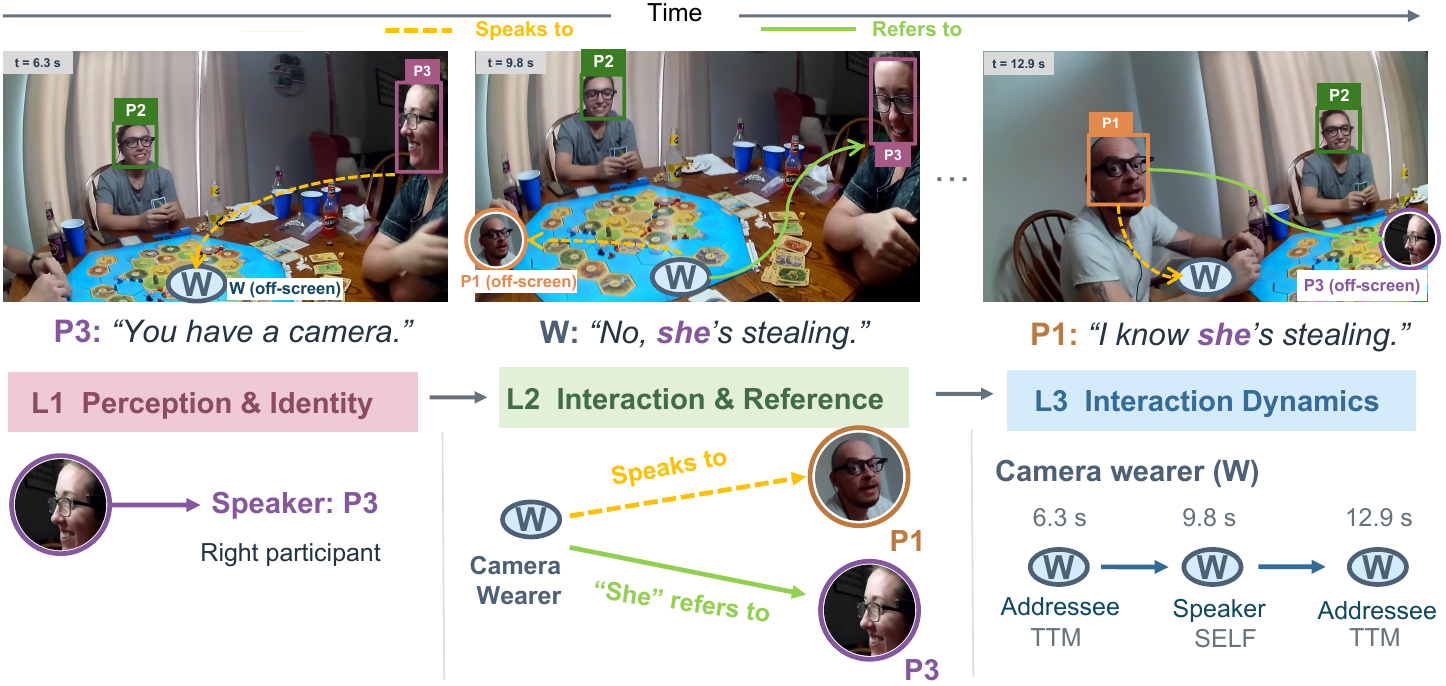}
    \vspace{-3mm}
    \caption{
        \textbf{Overview of MSIU-Bench.}
    The example illustrates three capability levels:
    \textbf{L1, social signal perception and identity grounding}:
    identifying P3 as the speaker;
    \textbf{L2, current social interaction and referential reasoning}:
    determining that the camera wearer W addresses P1 and that
    ``she'' refers to P3; and
    \textbf{L3, dynamic interaction-state reasoning}:
    tracking W's changing role from addressee to speaker and back
    to addressee.
    Yellow dashed arrows indicate who speaks to whom,
    and green solid arrows indicate referents.
    }
    \label{fig:msiu_overview}
    \vspace{-3mm}
\end{figure*}

\section{Introduction}
\label{sec:intro}

Understanding social interaction is fundamental to
intelligent assistants that respond appropriately
in everyday conversations~\cite{vinciarelli2009social, skantze2021turntaking}.
In multiparty dialogue, a speaker may address one
participant while referring to another, and participants
may shift between speaking, being addressed, and
not participating in the current exchange~\cite{inoue2025addressee}.
Egocentric settings introduce an additional challenge:
the camera wearer may be outside the field of view,
so their involvement must be inferred from the
surrounding speech and behavior~\cite{jiang2022egocentric}.
Such understanding requires multimodal large language
models (MLLMs) to associate visual, auditory, and
linguistic cues with specific participants~\cite{li2025tiktalkcoref}, interpret
their conversational relationships, and track how these
relationships evolve~\cite{jia2024avconv,
chang2026conversation,chowdhury2026amuse, ouyang2026multispeaker}.
Systematic evaluation of these abilities is therefore
an important step toward socially aware assistants.

Existing benchmarks evaluate multimodal models across
a range of video understanding and social interaction
tasks~\cite{xiao2021nextqa,patraucean2023perception, nguyen2026avspeakerbench, li2024mvbench, zadeh2019socialiq,kong2026sivbench}. However, further systematic evaluation is needed to
determine how reliably these models understand
participants' social roles and conversational
relationships, particularly how these relationships
change over time. Performance on an individual social task provides only a partial view of a model's capabilities and cannot
fully capture its strengths and weaknesses in
participant grounding, relational understanding,
and temporal reasoning. A unified benchmark is
therefore needed to systematically examine MLLMs'
capabilities and limitations in social interaction
understanding and identify key challenges for future
research.

We introduce \textbf{MSIU-Bench} (Multimodal Social
Interaction Understanding Benchmark), a dataset and
benchmark for multimodal social interaction understanding.
Through annotation alignment, task-specific question
and distractor design, and human review, we construct
40,417 question--answer (QA) pairs across 4,260 video clips.
The benchmark covers wearer-centered interactions and
multiparty conversations, drawing on existing videos
and annotations~\cite{du2026onlinettm,li2026onlinemmsi}. As illustrated in Fig.~\ref{fig:msiu_overview},
we organize 24 tasks into three capability levels: \emph{social signal perception and identity grounding}, \emph{current social interaction and referential reasoning}, and \emph{dynamic interaction-state reasoning}. We assess these capabilities through three complementary evaluation components.
The 12 \textbf{Core} tasks provide the primary evaluation,
requiring models to identify specific roles, participants,
and referents and to infer interaction states or transitions.
The 9 \textbf{Binary} tasks assess whether specific
social conditions hold, providing complementary evidence
for interpreting Core performance.
The 3 \textbf{Description-MCQ} tasks
(Description-based Multiple-Choice Questions) require
models to integrate information about the scene and
interaction and select the natural-language description
that best matches the observed interaction from four
options.

We systematically evaluate 21 MLLMs on MSIU-Bench, including 19 open-weight and two proprietary models. The strongest direct-inference baseline, Gemini-4-Argon~\cite{google2026gemini4argon}, achieves 66.35\% overall Core task-macro accuracy. Task- and class-level analyses reveal recurring difficulties in attributing utterances to the correct participants, recognizing non-participation, and identifying the direction of interaction-state changes. We further conduct supervised post-training on four open-weight models using LoRA~\cite{hu2022lora} and our training split, improving overall Core accuracy by 25.36--30.81 percentage points. Notably, D-ORCA-8B-0210~\cite{tang2026dorca} improves from 37.59\% to 62.95\%, approaching Gemini-4-Argon's 66.35\%. The adapted models also surpass Gemini-4-Argon on complementary evaluations, with D-ORCA-8B-0210 achieving 79.90\% versus 77.39\% on Binary and Gemma-4-12B-it achieving 86.62\% versus 75.55\% on Description-MCQ. These results demonstrate the value of our training data in enabling smaller open-weight models to approach or surpass frontier proprietary performance on specific social understanding tasks, while highlighting challenges that persist after adaptation.

Our main contributions are as follows:
\begin{itemize}
    \item We introduce \textbf{MSIU-Bench}, a dataset
    and benchmark containing 40,417 QA
    pairs across 24 tasks.
    Its three capability levels and complementary
    Core, Binary, and Description-MCQ components
    support systematic evaluation of multimodal
    social interaction understanding.

    \item We evaluate 21 MLLMs across model families and sizes. Analyses at the task and class levels reveal recurring difficulties in participant grounding, interaction-state recognition, and reasoning about changes in social interaction.

    \item We establish LoRA-based post-training baselines for four open-weight models, demonstrating the value of our training data in improving social interaction understanding while revealing persistent challenges.
\end{itemize}

\begin{figure*}[t]
    \centering
    \includegraphics[width=\linewidth]{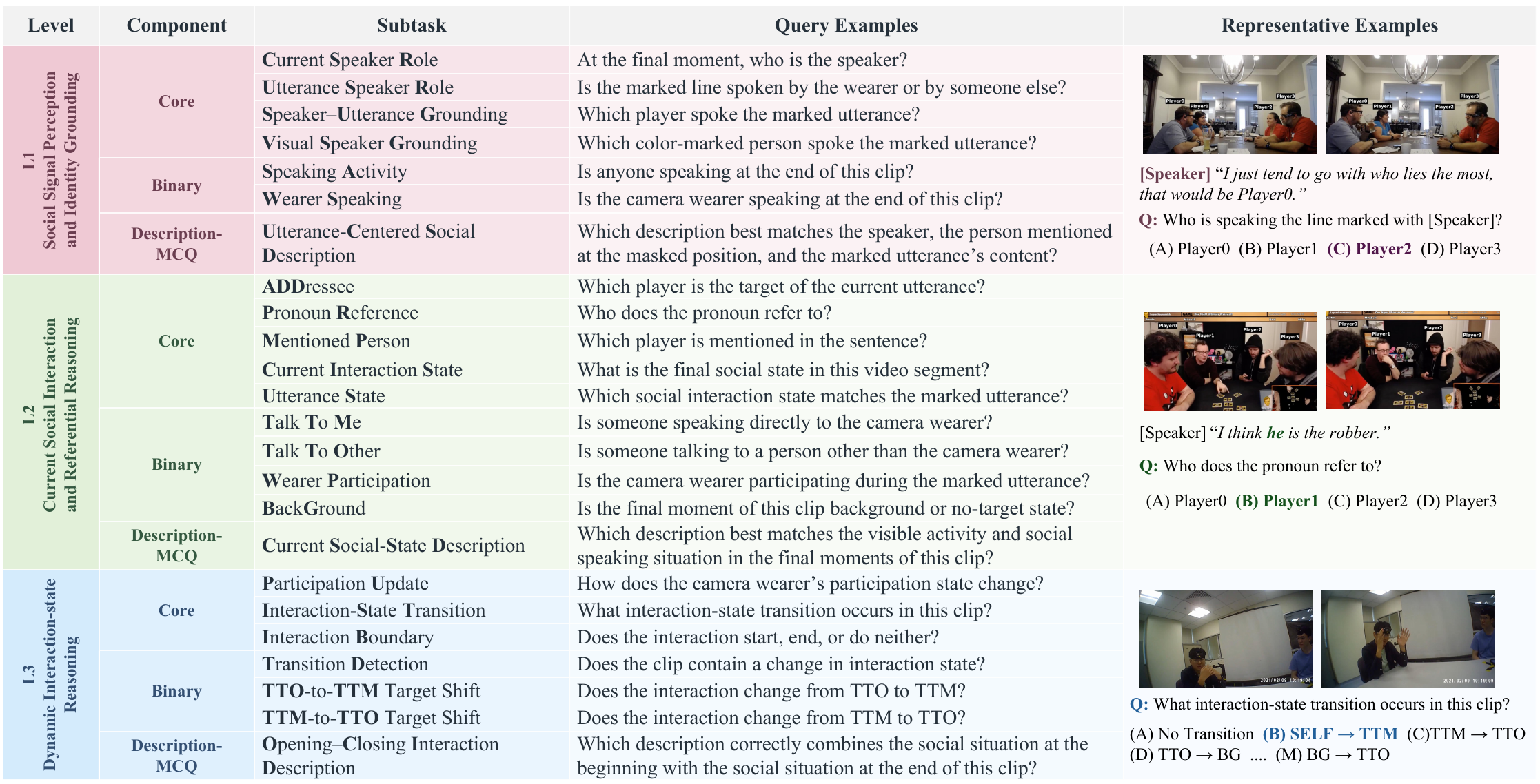}
    \vspace{-3mm}
    \caption{
\textbf{MSIU-Bench task overview.}
The 24 subtasks are grouped into three capability
levels, with representative examples.
}
    \label{fig:task_overview}
    \vspace{-3mm}
\end{figure*}

\section{Related Work}
\label{sec:related_work}

\noindent
\textbf{Multimodal Video Understanding Benchmarks.}
Video understanding benchmarks such as MVBench
and Video-MME~\cite{li2024mvbench,fu2025videomme}
evaluate multimodal models on perceptual
and reasoning tasks, including action recognition,
event understanding, and temporal
reasoning~\cite{zhou2025mlvu,wu2024longvideobench,
mangalam2023egoschema}.
Other benchmarks study embodied cognition and
egocentric memory~\cite{dang2025ecbench,ye2025mmego},
as well as audio-visual understanding in first-person
videos~\cite{seth2026egoavu,zhu2026egosound}. Different from general-purpose video understanding
benchmarks, we design MSIU-Bench to evaluate
participant-level social interaction understanding.

\vspace{1mm}
\noindent
\textbf{Social Interaction Understanding.}
Social-IQ and EmoBench-M evaluate social and emotional
understanding~\cite{zadeh2019socialiq,hu2026emobench}.
Related benchmarks examine reasoning about intentions
and mental states~\cite{li2023intentqa,jin2024mmtomqa,
kim2023fantom,chen2024tombench}.
Social Genome~\cite{mathur2025socialgenome} investigates
the evidence supporting social judgments, while
SIV-Bench and SMMU~\cite{kong2026sivbench,chen2026smmu}
extend evaluation to contextual social reasoning
and prediction. These benchmarks cover diverse aspects of social
understanding, with an emphasis on interpreting
mental states and the meaning or possible consequences
of interactions.

A more closely related line of work examines
wearer-directed speech and interaction states
in egocentric videos~\cite{grauman2022ego4d,
du2026onlinettm}.
MMSI and Online-MMSI~\cite{lee2024modeling,
li2026onlinemmsi} examine speaking targets and
conversational references, while
Omni-MMSI~\cite{li2026omnimmsi} explicitly evaluates
the attribution of audio-visual cues to participant
identities. Different from evaluations centered on state recognition
or participant reference resolution, we design MSIU-Bench
to systematically assess MLLMs across three capability
levels, characterizing their strengths and weaknesses
in grounding participants, reasoning about conversational
relations, and interpreting interaction-state changes.

\section{MSIU-Bench}
\label{sec:egosi_bench}

\subsection{Benchmark Overview}
\label{sec:benchmark_overview}

MSIU-Bench evaluates multimodal social interaction
understanding through video-based question answering.
Models receive a video clip, accompanying audio when
available, and any task-specific textual context.
The tasks span three capability levels and are
assessed through Core, Binary, and Description-MCQ
(Fig.~\ref{fig:task_overview}).
We align source annotations to construct task-specific
questions and answer options, then apply automated
checks and human review.
The resulting dataset provides training, validation,
and test splits for model adaptation and evaluation.

\subsection{Task Taxonomy}
\label{sec:task_taxonomy}

Understanding changes in social interaction involves
several related judgments. For example, a person may
initially speak to someone else and then begin addressing
the camera wearer. Understanding this change involves
identifying the speaker, inferring the addressee, and
comparing interaction relations across time.

Motivated by these requirements, we organize the capability
framework of MSIU-Bench into three levels:
\emph{social signal perception and identity grounding},
\emph{current social interaction and referential reasoning},
and \emph{dynamic interaction-state reasoning}.
Across 24 tasks, we assess both specific social judgments
and the joint understanding of related social facts through
natural-language descriptions.
Individual tasks may involve one or more capability levels.
The three levels address three central questions:
\emph{Who is speaking?},
\emph{Who is interacting with whom?}, and
\emph{How do interaction states and participation
change over time?}
This organization supports analysis of model performance
in perception, identity grounding, referential reasoning,
and temporal social reasoning.
Figure~\ref{fig:task_overview} illustrates the three levels
with representative questions and summarizes all 24 tasks.

These three levels capture distinct but complementary capabilities required for multimodal social interaction understanding. L1 grounds speech and utterances in participant identities, L2 establishes the conversational relationships among these participants, and L3 reasons about how their roles and relationships change over time. Success at one level does not guarantee success at another: identifying a speaker does not establish whom they address, and recognizing interaction states at individual moments does not necessarily reveal how those states change. Evaluating the three levels separately allows us to distinguish weaknesses in participant grounding, relational reasoning, and temporal social understanding.

\vspace{1mm}
\noindent\textbf{L1: Social Signal Perception and Identity
Grounding.}
This level asks \emph{who is speaking and which participant
a speech signal or utterance belongs to}.
These tasks
ask models to detect speech, identify the speaker's role or
identity, and associate a marked utterance with the
corresponding participant in the dialogue or visual scene.
These judgments establish the participants associated with
the observed signals, providing the basis for interpreting
their interactions. For example, in Fig.~\ref{fig:msiu_overview}, identifying P3 as
the speaker grounds the first utterance in the participant
who produced it.

\begin{figure*}[t]
    \centering
    \includegraphics[width=\linewidth]{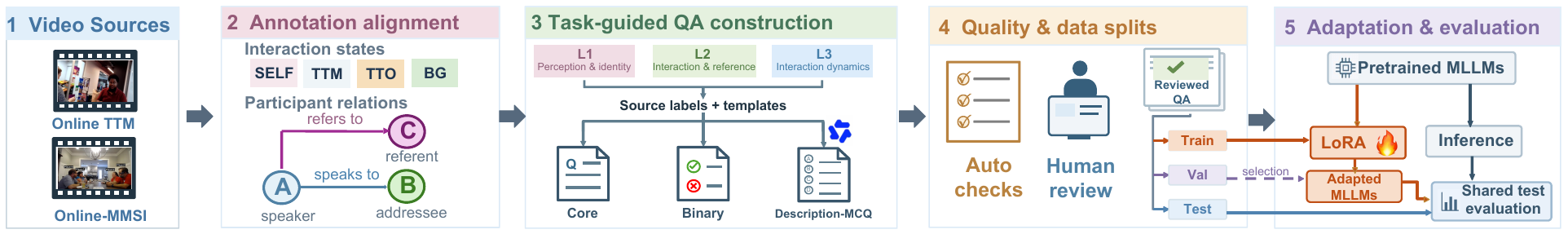}
    \vspace{-7mm}
    \caption{\textbf{Construction and evaluation pipeline of MSIU-Bench.}
    We align source annotations and construct task-specific QA pairs across three capability levels, followed by automated checks, human review, and data splitting. The resulting splits support LoRA adaptation, model selection, and evaluation of pretrained and adapted MLLMs on a shared test set.}
\label{fig:msiu_pipeline}
\vspace{-5mm}
\end{figure*}

\vspace{1mm}
\noindent\textbf{L2: Current Social Interaction and Referential Reasoning.}
This level asks \emph{who is addressing whom, who is being
referred to, and how participants are involved}.
Tasks assess addressees, pronoun referents, mentioned
participants, wearer participation, and interaction states
at a specified moment or utterance.
The distinction between these roles is essential: an
utterance can address one participant while referring
to another. For example, in Fig.~\ref{fig:msiu_overview}, the camera wearer
addresses P1 while using ``she'' to refer to P3. Correctly
identifying the speaker is therefore insufficient: the model
must distinguish the addressee from the referent.

\vspace{1mm}
\noindent
\textbf{L3: Dynamic Interaction-State Reasoning.}
This level asks \emph{how participation and interaction
relations change over time}.
Tasks assess participation updates, ordered state
transitions, interaction boundary understanding, and directed
changes in speaking targets.
For example, a shift from addressing another person to
addressing the camera wearer changes the wearer's
participation even though both moments involve non-wearer
speech. Recognizing speech at both moments is insufficient:
the model must compare the interaction targets and preserve
their temporal order. For example, the camera wearer in Fig.~\ref{fig:msiu_overview}
changes from addressee to speaker and back to addressee.
A single-time judgment cannot capture this sequence.
The transition tasks distinguish
\texttt{TTO} (\emph{talk to others})~$\rightarrow$~
\texttt{TTM} (\emph{talk to me}) from the reverse
transition; ``me'' refers to the camera wearer. The state-transition
task includes the 12 directed transitions between distinct
states and a \texttt{No Transition} class.

This organization distinguishes failures in participant grounding, relational interpretation, and temporal tracking. Tasks are grouped by their primary capability while recognizing that they may draw on shared evidence and involve multiple capability levels.

\subsection{Evaluation Design}
\label{sec:evaluation_components}

The three capability levels characterize \emph{what} social information models must understand, while the evaluation components assess different aspects of this understanding. MSIU-Bench evaluates not only whether models recognize social signals, but also whether they ground these signals in participants, interpret changing interaction relationships, and integrate multiple social facts. We therefore organize the 24 tasks into three complementary components: Core, Binary, and Description-MCQ.

\noindent
\textbf{Core}
comprises 12 tasks requiring a specific role, participant,
referent, interaction state, or transition category.
It provides the primary assessment of whether models
can identify the correct social interpretation.

\noindent
\textbf{Binary}
comprises nine tasks that assess specific social conditions,
providing focused diagnostics for related Core tasks.
For example, while transition classification identifies
the specific state change, directional Binary questions
enable separate analysis of missed target shifts and
false alarms in each direction.

\noindent
\textbf{Description-MCQ}
comprises three tasks that ask models to select the
description that best matches the observed interaction
from four options. The tasks cover participant--utterance relations,
current interaction states, and opening--closing
states. For the last task, only the two endpoint
states and their temporal order are assessed.

We report task-macro accuracy separately for each
component, assigning equal weight to its constituent
tasks:
\begin{equation}
    \mathrm{Acc}_{c}
    =
    \frac{1}{|\mathcal{T}_{c}|}
    \sum_{t \in \mathcal{T}_{c}} \mathrm{Acc}_{t},
    \label{eq:task_macro}
\end{equation}
where $c$ denotes an evaluation component,
$\mathcal{T}_{c}$ is its task set, and
$\mathrm{Acc}_{t}$ is the test accuracy on task $t$.
Each task has equal weight, regardless of its number
of questions. Core task-macro accuracy is the primary
metric. Level-specific scores are computed in the
same way over tasks in each level.

\begin{table}[t]
\centering
\caption{
\textbf{Comparison of social interaction data sources.}
Ego: human-worn egocentric video; Multi: multi-person scenes;
V/A/T: video/audio/dialogue text.
Frame: frame-level activity or state labels;
4-State: unified SELF/TTM/TTO/BG sequences;
Target: speech addressees; Ref.: participant references.
}
 \vspace{-3mm}
\begingroup
\small
\setlength{\tabcolsep}{4.5pt}
\renewcommand{\arraystretch}{1.08}
\scalebox{0.75}{
\begin{tabular}{@{}l cc ccc cccc@{}}
\toprule

& \multicolumn{2}{c}{\textbf{Setting}}
& \multicolumn{3}{c}{\textbf{Modalities}}
& \multicolumn{4}{c}{\textbf{Annotations}} \\

\cmidrule(lr){2-3}
\cmidrule(lr){4-6}
\cmidrule(l){7-10}

\textbf{Dataset}
& \textbf{Ego}
& \textbf{Multi}
& \textbf{V}
& \textbf{A}
& \textbf{T}
& \textbf{Frame}
& \textbf{4-State}
& \textbf{Target}
& \textbf{Ref.} \\
\midrule

SocialDial~\cite{zhan2023social}
& \msiuno & \msiuno
& \msiuno & \msiuno & \msiuyes
& \msiuno & \msiuno & \msiuno & \msiuno \\

JRDB-Social~\cite{jahangard2024jrdb}
& \msiuno & \msiuyes
& \msiuyes & \msiuyes & \msiuno
& \msiuyes & \msiuno & \msiuno & \msiuno \\

EasyCom~\cite{donley2021easycom}
& \msiuyes & \msiuyes
& \msiuyes & \msiuyes & \msiuyes
& \msiuyes & \msiuno & \msiuyes & \msiuno \\

Ego4D Social~\cite{grauman2022ego4d}
& \msiuyes & \msiuyes
& \msiuyes & \msiuyes & \msiuyes
& \msiuyes & \msiuno & \msiuyes & \msiuno \\

MMSI~\cite{lee2024modeling}
& \msiuno & \msiuyes
& \msiuyes & \msiuyes & \msiuyes
& \msiuno & \msiuno & \msiuyes & \msiuyes \\

\midrule

\rowcolor{blue!5}
\textbf{Online TTM}~\cite{du2026onlinettm}
& \msiuyes & \msiuyes
& \msiuyes & \msiuyes & \msiuyes
& \msiuyes & \msiuyes & \msiuyes & \msiuno \\

\rowcolor{blue!5}
\textbf{Online-MMSI}~\cite{li2026onlinemmsi}
& \msiuno & \msiuyes
& \msiuyes & \msiuyes & \msiuyes
& \msiuno & \msiuno & \msiuyes & \msiuyes \\

\bottomrule
\end{tabular}
}
\endgroup
\label{tab:source_comparison}
\vspace{-3mm}
\end{table}

\subsection{Benchmark Construction}
\label{sec:qa_construction}

Figure~\ref{fig:msiu_pipeline} summarizes the construction
and evaluation pipeline. We build QA instances from
existing temporal and conversational annotations,
with answers anchored in the corresponding source evidence. We convert these annotations into task-specific
questions and answer choices covering participant
grounding, interaction relations, and ordered state
changes. The Core, Binary, and Description-MCQ
formats provide complementary views of model
performance on these social judgments.

\noindent
\textbf{Complementary data sources.} MSIU-Bench draws on two complementary sources that provide
different forms of social interaction supervision
(Table~\ref{tab:source_comparison}).

\emph{Online TTM}~\cite{du2026onlinettm} builds on Ego4D
Social annotations~\cite{grauman2022ego4d} and provides
frame-level interaction states from the camera wearer's
perspective. Its four states distinguish wearer speech
(\texttt{SELF}), speech directed toward the wearer
(\texttt{TTM}), speech directed toward another person
(\texttt{TTO}), and background conditions (\texttt{BG}).
These annotations support questions about speaking roles,
wearer participation, interaction states, and their
changes over time.

\emph{Online-MMSI}~\cite{li2026onlinemmsi} follows the
participant-reference tasks introduced in
MMSI~\cite{lee2024modeling}. The resources build on the
Ego4D and YouTube portions of Werewolf Among
Us~\cite{lai2023werewolf}, using recordings of multiparty
social deduction games. They associate dialogue with
participant identities and provide annotations for
speaking target identification, pronoun coreference
resolution, and mentioned player prediction.
We use these annotations to construct questions about
speakers, addressees, and referenced participants. Together, the two sources enable us to assess both
wearer-centered interaction states and participant-level
conversational relations.

\noindent\textbf{Annotation-grounded QA construction.}
We align temporal annotations with clip endpoints,
transcribed utterances, and annotated transition events
to derive task-specific answers.
For multiparty dialogue, source speaker identities,
participant references, and visual identity markers
support grounding questions.
Each question targets a defined social judgment.
A video clip may be used to construct multiple
questions across the three evaluation components.
Source and temporal metadata are retained for verification.

For Description-MCQ, we combine structured social
annotations with scene information from sampled frames.
We use Qwen3-VL-2B-Instruct~\cite{bai2025qwen3vl}
to generate visual descriptions for the initial
candidates, and Qwen2.5-Omni-7B~\cite{xu2025qwen25omni}
for subsequent description rewriting and dataset
expansion.
Speaker identities, participant relations, and
interaction states remain constrained by the
source annotations.
Each question contains one candidate correct description
and three distractors that alter information such as
speaker attribution, participant relations, or temporal order.

\noindent
\textbf{Quality control.}
Automated checks examine required fields, video paths,
and answer formats. All 40,417 QA instances undergo
human review to check whether the labeled answers
agree with the available video, audio, and context.
For Description-MCQ, reviewers additionally assess
whether the labeled answer is uniquely supported and
record errors, ambiguities, and suggested corrections.

\begin{table}[t]
    \centering
    \caption{\textbf{QA counts by evaluation component and data split.}}
    \vspace{-3mm}
    \label{tab:dataset_splits}
    \small
    \resizebox{0.98\columnwidth}{!}{%
    \begin{tabular}{lrrrr}
        \toprule
        Component & Train & Validation & Test & Total \\
        \midrule
        Core
        & 15,860 & 2,218 & 4,545 & 22,623 \\
        Binary
        & 9,696 & 1,352 & 2,782 & 13,830 \\
        Description-MCQ
        & 2,776 & 399 & 789 & 3,964 \\
        \midrule
        \textbf{Overall}
        & \textbf{28,332}
        & \textbf{3,969}
        & \textbf{8,116}
        & \textbf{40,417} \\
        \bottomrule
    \end{tabular}%
    }
    \vspace{-5mm}
\end{table}

\begin{table*}[t]
\centering
\caption{
\textbf{Results on MSIU-Bench.}
We report task-macro accuracy (\%) by capability
level and evaluation component. D-MCQ denotes
Description-MCQ. \textbf{Bold} and
\underline{underlined} scores indicate the best and
second-best results across all models, respectively.
}
\small
\setlength{\tabcolsep}{3pt}
\renewcommand{\arraystretch}{0.95}
\setlength{\aboverulesep}{1pt}
\setlength{\belowrulesep}{1pt}

\resizebox{\textwidth}{!}{%
\begin{tabular}{@{}l@{\hspace{0.9em}}rrr@{\hspace{0.9em}}rrr@{\hspace{0.9em}}rrr@{\hspace{0.9em}}rrr@{}}
\toprule
& \multicolumn{3}{c}{\textit{Level 1}}
& \multicolumn{3}{c}{\textit{Level 2}}
& \multicolumn{3}{c}{\textit{Level 3}}
& \multicolumn{3}{c}{\textit{Overall}} \\
\cmidrule(lr){2-4}
\cmidrule(lr){5-7}
\cmidrule(lr){8-10}
\cmidrule(lr){11-13}
\textbf{Model}
 & Core & Binary & D-MCQ
 & Core & Binary & D-MCQ
 & Core & Binary & D-MCQ
 & Core & Binary & D-MCQ \\
\midrule

\multicolumn{13}{c}{\textit{Closed-source Models}} \\
\midrule
Gemini-4-Argon~\cite{google2026gemini4argon}
 & \textbf{70.74} & 83.06 & 88.81
 & \textbf{74.47} & \underline{85.65} & 71.91
 & 46.94 & 62.59 & 65.93
 & \textbf{66.35} & 77.39 & 75.55 \\
Gemini-3.5-Flash-Lite~\cite{google2026gemini35flashlite}
 & 56.62 & 80.62 & 75.42
 & 53.48 & 73.23 & 65.43
 & 20.53 & 49.28 & 48.42
 & 46.29 & 66.89 & 63.09 \\

\midrule
\multicolumn{13}{c}{\textit{Model Size $\geq 14$B}} \\
\midrule
Qwen3-Omni-30B-A3B-Instruct~\cite{xu2025qwen3omni}
 & 40.00 & 56.84 & 69.46
 & 47.39 & 54.58 & 69.66
 & 13.69 & 46.52 & 47.25
 & 36.50 & 52.40 & 62.13 \\
Nemotron-3-Nano-Omni-30B-A3B~\cite{nvidia2026nemotronnanoomni}
 & 28.59 & 72.15 & 81.82
 & 32.29 & 55.11 & 64.61
 & 20.61 & 49.15 & 50.00
 & 28.14 & 56.91 & 65.47 \\
video-SALMONN 2 Pro 32B~\cite{tang2025videosalmonn2,videosalmonn2pro32b}
 & 54.46 & 83.88 & 56.64
 & 52.48 & 64.91 & 43.82
 & 22.66 & 61.24 & 44.51
 & 45.69 & 67.90 & 48.32 \\
Ming-Lite-Omni-1.5~\cite{inclusionai2025mingomni}
 & 35.46 & 56.51 & 68.53
 & 47.53 & 59.16 & 47.19
 & 20.44 & 52.90 & 56.04
 & 36.74 & 56.48 & 57.26 \\
Qwen3-Omni-30B-A3B-Thinking~\cite{xu2025qwen3omni}
 & 47.07 & 72.80 & 76.69
 & 53.25 & 68.79 & 61.80
 & 31.05 & 58.41 & 50.00
 & 45.64 & 66.22 & 62.83 \\

\midrule
\multicolumn{13}{c}{\textit{Model Size: $7\mathrm{B} \leq \mathrm{Size} < 14\mathrm{B}$}} \\
\midrule
D-ORCA-8B-0210~\cite{tang2026dorca}
 & 47.84 & 76.22 & 58.74
 & 41.69 & 70.54 & 55.06
 & 17.08 & 60.53 & 45.05
 & 37.59 & 68.47 & 52.95 \\
Qwen2.5-Omni-7B~\cite{xu2025qwen25omni}
 & 18.00 & 67.43 & 65.73
 & 40.31 & 51.72 & 55.06
 & 23.36 & 51.31 & 36.26
 & 28.64 & 55.07 & 52.35 \\
MiniCPM-o 4.5 (9B)~\cite{cui2026minicpmo45}
 & 38.89 & 66.29 & 68.76
 & 46.41 & 69.73 & 50.56
 & 20.54 & 60.45 & 29.67
 & 37.44 & 65.87 & 49.67 \\
Ola-7B~\cite{liu2025ola}
 & 25.86 & 63.03 & 66.43
 & 40.40 & 55.89 & 39.89
 & 17.66 & 56.64 & 38.46
 & 29.87 & 57.73 & 48.26 \\
video-SALMONN 2+ 7B~\cite{tang2025videosalmonn2}
 & 16.54 & 53.42 & 59.21
 & 36.79 & 51.23 & 38.76
 & 29.06 & 52.08 & 36.26
 & 28.11 & 52.00 & 44.75 \\
HumanOmni-7B~\cite{zhao2025humanomni}
 & 34.50 & 55.54 & 56.18
 & 40.84 & 46.83 & 43.26
 & 9.77 & 41.87 & 34.07
 & 30.96 & 47.11 & 44.50 \\
VideoLLaMA2.1-7B-AV~\cite{cheng2024videollama2}
 & 18.11 & 62.54 & 56.64
 & 39.02 & 48.49 & 38.76
 & 32.54 & 42.17 & 34.62
 & 30.43 & 49.51 & 43.34 \\
OmniVinci-9B~\cite{ye2025omnivinci}
 & 35.51 & 38.60 & 56.18
 & 35.29 & 47.31 & 36.52
 & 24.04 & 41.56 & 36.81
 & 32.55 & 43.46 & 43.17 \\
Gemma-4-12B-it~\cite{gemmateam2026gemma4}
 & 25.79 & 52.93 & 77.86
 & 27.72 & 57.34 & 60.11
 & 32.94 & 58.55 & 41.76
 & 28.38 & 56.76 & 59.91 \\

\midrule
\multicolumn{13}{c}{\textit{Model Size $< 7$B}} \\
\midrule
Qwen2.5-Omni-3B~\cite{xu2025qwen25omni}
 & 19.28 & 42.83 & 68.07
 & 26.28 & 49.64 & 53.93
 & 27.96 & 46.54 & 34.07
 & 24.36 & 47.09 & 52.02 \\
video-SALMONN 2+ 3B~\cite{tang2025videosalmonn2}
 & 37.23 & 41.53 & 62.70
 & 31.53 & 48.51 & 38.76
 & 24.91 & 53.61 & 31.32
 & 31.77 & 48.66 & 44.26 \\
video-SALMONN 2 Pro 4B~\cite{tang2025videosalmonn2,videosalmonn2pro4b}
 & 36.77 & 62.54 & 73.43
 & 48.31 & 59.10 & 58.99
 & 37.91 & 60.01 & 56.04
 & 41.86 & 60.17 & 62.82 \\
Phi-4-multimodal-instruct (5.6B)~\cite{abouelenin2025phi4mini}
 & 30.11 & 57.62 & 68.32
 & 31.20 & 54.30 & 49.50
 & 31.67 & 57.60 & 45.53
 & 30.95 & 56.14 & 54.45 \\
Gemma-4-E2B-it (5.1B)~\cite{gemmateam2026gemma4}
 & 32.22 & 39.74 & 62.24
 & 23.86 & 57.70 & 62.92
 & 22.18 & 58.11 & 28.57
 & 26.23 & 53.84 & 51.24 \\

\midrule
\multicolumn{13}{c}{\textit{LoRA Post-training}~\cite{hu2022lora}} \\
\midrule
\textbf{Qwen2.5-Omni-7B + LoRA}
 & 60.31 & \underline{94.14} & \underline{93.07}
 & 58.44 & 83.58 & 85.31
 & 57.68 & \underline{62.87} & 61.98
 & 58.87 & 79.02 & 80.12 \\
\textbf{Qwen2.5-Omni-3B + LoRA}
 & 58.61 & \textbf{94.46} & 85.31
 & 52.01 & 84.47 & 80.45
 & 55.60 & 54.45 & 49.13
 & 55.11 & 76.68 & 71.63 \\
\textbf{D-ORCA-8B-0210 + LoRA}
 & \underline{63.11} & 93.49 & 92.77
 & \underline{60.75} & \textbf{86.66} & \underline{87.44}
 & \underline{66.40} & 61.84 & \underline{67.68}
 & \underline{62.95} & \textbf{79.90} & \underline{82.63} \\
\textbf{Gemma-4-12B-it + LoRA}
 & 56.20 & 93.62 & \textbf{94.65}
 & 54.64 & 84.60 & \textbf{89.06}
 & \textbf{70.74} & \textbf{63.94} & \textbf{76.15}
 & 59.19 & \underline{79.72} & \textbf{86.62} \\

\bottomrule
\end{tabular}%
}
\label{tab:msiu-level-overall}
\vspace{-5mm}
\end{table*}

\subsection{Dataset Statistics and Splits}
\label{sec:dataset_statistics}

\noindent\textbf{Dataset statistics.}
MSIU-Bench contains 40,417 QA instances across 24 tasks.
Online TTM and Online-MMSI contribute 31,465 and 8,952
instances, respectively.
The dataset references 4,260 distinct video clips, with a cumulative duration of approximately 22.13 hours.
Grouped by each task's primary capability level,
L1, L2, and L3 contain 15,510, 18,087, and 6,820 QA instances,
respectively.
Across evaluation components, Core, Binary, and Description-MCQ
contain 22,623, 13,830, and 3,964 instances, respectively.

\noindent\textbf{Dataset splits.}
To support model adaptation and standardized evaluation,
we partition the dataset into training, validation, and test
sets at an approximate ratio of 7:1:2,
as summarized in Table~\ref{tab:dataset_splits}.
All three splits cover all 24 tasks and are disjoint
at the video-clip level. Questions derived from the
same clip, including questions from different
evaluation components, are assigned to the same split. The training split provides supervision for model adaptation,
including LoRA fine-tuning, allowing us to investigate the
dataset's utility for improving social understanding.
The validation split supports configuration tuning and model
selection during development, keeping these decisions separate
from final evaluation.
The test split contains 8,116 QA instances and serves as the
fixed evaluation set for model comparison.

\section{Experiments}
\label{sec:experiments}

\subsection{Experimental Setup}
\label{sec:experimental_setup}

\textbf{Models.}
We evaluate 21 multimodal large language models on the
fixed test set without task-specific fine-tuning.
The models span a range of families and sizes.
The 19 open-weight models include members of the
Qwen-Omni~\cite{xu2025qwen25omni,xu2025qwen3omni},
video-SALMONN~\cite{tang2025videosalmonn2, videosalmonn2pro32b,videosalmonn2pro4b},
and Gemma 4~\cite{gemmateam2026gemma4} families,
as well as D-ORCA~\cite{tang2026dorca},
HumanOmni~\cite{zhao2025humanomni},
and OmniVinci~\cite{ye2025omnivinci}.
The two closed-source models are
Gemini-3.5-Flash-Lite~\cite{google2026gemini35flashlite}
and Gemini-4-Argon~\cite{google2026gemini4argon}.
Table~\ref{tab:msiu-level-overall} reports the results
for each model.

\noindent
\textbf{Model inputs and inference.}
For each test question, we provide the model with the video, question, answer choices, and any task-specific textual context. Audio is included according to each model's evaluation configuration. To ensure consistent evaluation, we keep the answer choice order and use zero-shot inference without in-context examples. Inputs are formatted according to each model's audio-visual interface. Additional implementation details, including frame sampling, spatial resolution, audio preprocessing, and decoding configurations, are provided in the appendix.

\noindent
\textbf{Evaluation protocol.}
We use the test set defined in
Sec.~\ref{sec:dataset_statistics}, which contains
8,116 question--answer pairs. We report accuracy separately for Core, Binary, and Description-MCQ, averaging task accuracies with equal weight within each component. Core accuracy is our primary metric
(Sec.~\ref{sec:evaluation_components}). For the structured tasks (Core and Binary), we also
report balanced accuracy~\cite{brodersen2010balanced}
and macro-F1~\cite{opitz2024closer}
to assess performance under class imbalance. We examine class-specific recall and false positives to identify which cases models miss and which they incorrectly flag.

\subsection{Benchmark Results and Analysis}
\label{sec:benchmark_results}

\begin{figure}[t]
    \centering
    \includegraphics[width=0.6\linewidth]{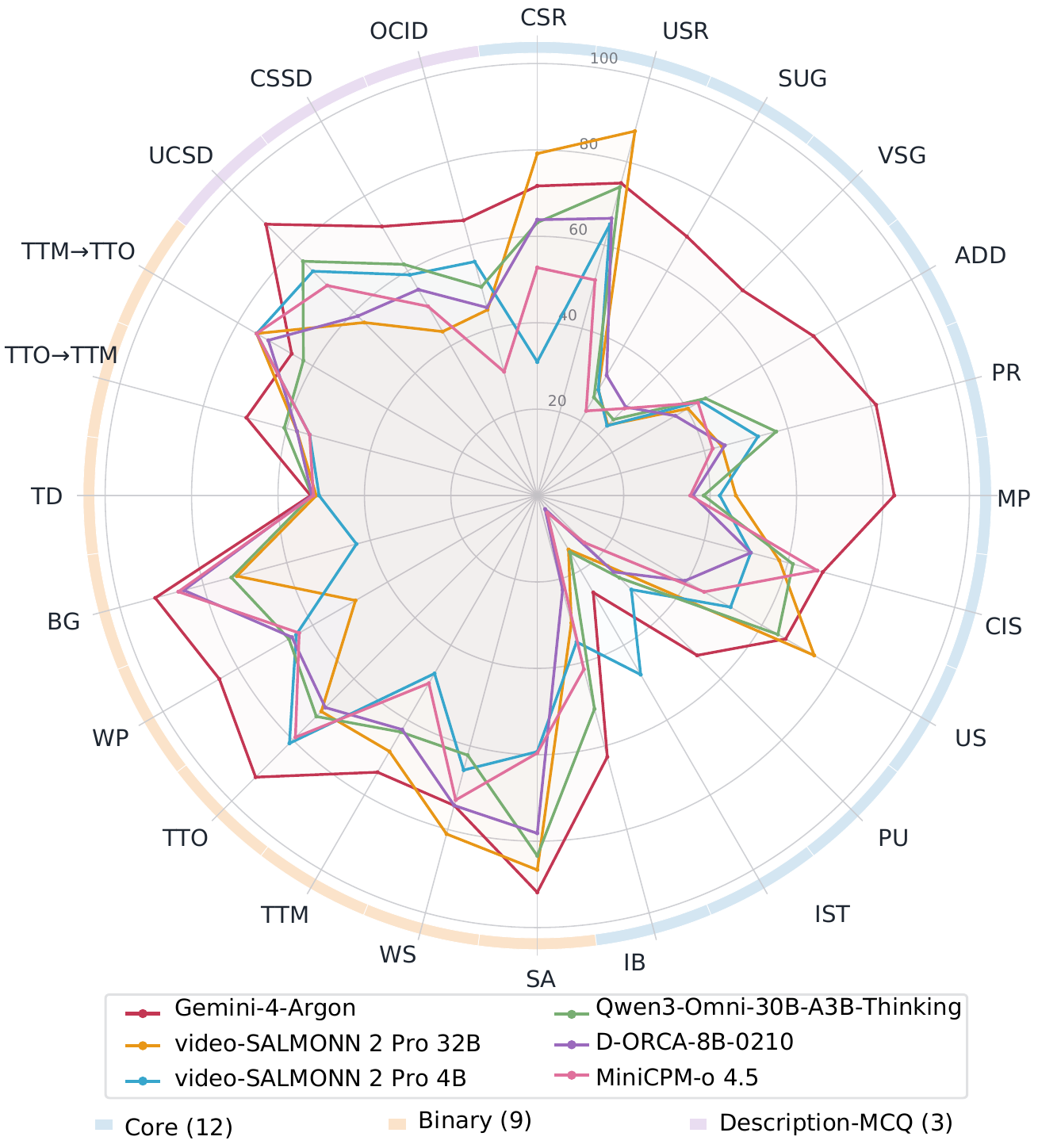}
    \vspace{-3mm}
    \caption{\textbf{Task-level accuracy (\%) of six direct-inference baselines on MSIU-Bench.}
    Each spoke represents a subtask; the outer ring groups
Core, Binary, and Description-MCQ.}
    \label{fig:baseline_subtasks}
    \vspace{-5mm}
\end{figure}

Table~\ref{tab:msiu-level-overall} reports performance across the three capability levels (L1--L3) and evaluation components. Figure~\ref{fig:baseline_subtasks} complements these aggregate
results with task-level accuracy for six direct-inference
baselines. We analyze the results at the level, task, and class levels to understand not only how well current MLLMs perform, but also \emph{which aspects of social interaction understanding remain difficult}. 

\vspace{1mm}
\noindent
\noindent\textbf{Even frontier MLLMs struggle with dynamic social interaction understanding.} Across 21 direct-inference baselines, substantial performance
gaps remain on MSIU-Bench. Among open-weight direct-inference baselines, video-SALMONN 2 Pro 32B achieves the highest overall Core
accuracy of 45.69\%. However, larger models do not
consistently outperform smaller counterparts, particularly
on dynamic interaction reasoning. Within the video-SALMONN 2
Pro family, increasing model size from 4B to 32B improves
overall Core accuracy from 41.86\% to 45.69\%, but decreases
L3 Core accuracy from 37.91\% to 22.66\%.
Gemini-4-Argon, a highly capable frontier MLLM, substantially
outperforms all open-weight baselines, achieving 66.35\%
overall Core accuracy, 20.66 percentage points above
the strongest open-weight model. Despite this strong
performance, its L3 Core accuracy reaches only 46.94\%,
highlighting persistent challenges in reasoning about
how social interactions evolve over time.
Together, these results reveal three key findings:
(1) open-weight MLLMs have substantial room for improvement
in social interaction understanding;
(2) increasing model size and advances in general multimodal
capabilities do not necessarily translate into stronger
dynamic social reasoning; and
(3) even highly capable frontier MLLMs exhibit significant
limitations in understanding how social interactions
evolve over time.


\vspace{1mm}
\noindent
\textbf{L1: Speech detection does not guarantee reliable speaker grounding.}
For example, the video-SALMONN 2 Pro 32B model achieves 86.64\% accuracy on speech activity detection. Its performance on current speaker-role identification, however, varies sharply by role. Although its accuracy exceeds that of Gemini-4-Argon (79.15\% vs.\ 71.66\%), it identifies the wearer as the speaker in only one of 18 wearer examples (5.56\% recall), whereas Argon identifies 16 of 18 (88.89\%). Higher task accuracy can thus conceal missed instances of the wearer speaking.

When asked to link an utterance to a specific participant, video-SALMONN 2 Pro 32B achieves 28.28\% on utterance grounding and 23.08\% on visual speaker grounding. Gemini-4-Argon reaches 69.23\% and 67.19\%, respectively. These results highlight a gap between detecting speech and identifying its source: {open-weight MLLMs can recognize speaking activity but struggle to associate utterances with the correct participants and their visual identities.}

\noindent
\textbf{L2: Models resolve references to people but struggle to recognize when the wearer is not participating.}
Gemini-4-Argon averages 74.47\% accuracy across the five
Level 2 Core tasks (Table~\ref{tab:msiu-level-overall}), reaching 81.16\% on pronoun resolution and 82.61\% on identifying the person mentioned. It achieves 84.95\% accuracy in judging whether the wearer participates in the interaction associated with a given utterance. The errors are asymmetric: it incorrectly predicts non-participation for only 4.26\% of examples in which the wearer participates, but predicts participation in 101 of the 150 examples in which the wearer does not (67.33\%).

Gemini-4-Argon and video-SALMONN 2 Pro 32B also miss different states when classifying the interaction at the end of a clip. For background states or those with no clear addressee (BG), Argon achieves 70.45\% recall, compared with 9.09\% for Pro 32B. Their relative performance reverses when one non-wearer explicitly addresses another (TTO): recall is 42.86\% for Argon and 64.29\% for Pro 32B. Pro 32B thus misses background states more often, whereas Argon more often misses speech directed from one non-wearer to another. These contrasting errors reveal model-specific
weaknesses in distinguishing background conditions from
interactions between non-wearer participants, highlighting
the importance of class-level evaluation in uncovering
limitations obscured by aggregate accuracy.

\noindent
\textbf{L3: Models recognize some speaker-role changes but often miss shifts in addressee.}
On interaction-state change detection, balanced accuracy across the direct-inference baselines ranges from 48.20\% to 53.80\%, close to the 50\% chance level for a binary decision. When asked to identify the type of transition, Gemini-4-Argon achieves 25.90\% accuracy. For switches between the wearer speaking (SELF) and another person addressing the wearer (TTM), its recall is 58.82\% for \(\mathrm{SELF}\to\mathrm{TTM}\) and 64.71\% for \(\mathrm{TTM}\to\mathrm{SELF}\). Recall falls to 25.00\% when the addressee shifts from someone else to the wearer (\(\mathrm{TTO}\to\mathrm{TTM}\)) and to 13.67\% for the reverse shift (\(\mathrm{TTM}\to\mathrm{TTO}\)). In these last two transitions, the wearer is neither the initial nor the final speaker. Distinguishing the directions requires tracking whom the speech is addressed to at each point.

A similar asymmetry appears in tracking changes in the wearer's participation. Gemini-4-Argon achieves 84.21\% recall when the wearer participates at both time points, compared with 35.21\% when the wearer joins and 31.11\% when the wearer leaves. The sharpest contrast is between the two cases in which participation does not change: recall is just 3.33\% when the wearer participates at neither time point. The model is most reliable when participation continues; it struggles to recognize entry, exit, and continued non-participation.

\vspace{1mm}
\noindent
\noindent\textbf{Complementary task formats reveal distinct transition reasoning failures.}
D-ORCA-8B-0210 achieves only 3.61\% accuracy on the
13-way Core interaction-state transition task, indicating
difficulty in identifying specific state changes.
Binary tasks provide finer-grained diagnostics:
D-ORCA achieves 57.58\% accuracy on
\(\mathrm{TTO}\to\mathrm{TTM}\) and 71.88\% on
\(\mathrm{TTM}\to\mathrm{TTO}\).
For the former, it detects 75.00\% of actual shifts
but produces 52.38\% false positives; for the latter,
recall drops to 58.33\%, while the false-positive rate
decreases to 20.00\%. These results reveal
direction-dependent errors obscured by aggregate accuracy.

\begin{figure}[t]
    \centering
    \includegraphics[width=\linewidth]{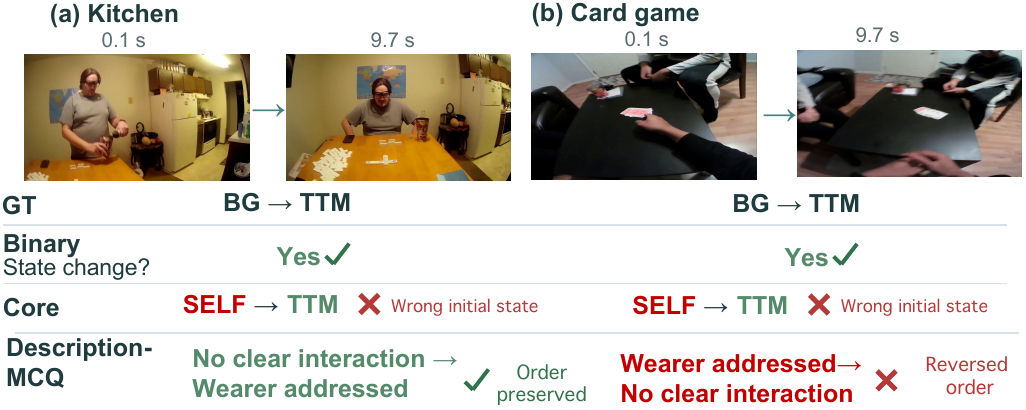}
    \vspace{-5mm}
    \caption{
        \textbf{Qualitative examples of interaction-state reasoning.}
        MiniCPM-o 4.5 predictions on a kitchen clip and a
        card-game clip. Description-MCQ text summarizes
        the event order in the model-selected options.
    }
    \label{fig:qualitative_cases}
    \vspace{-5mm}
\end{figure}

Description-MCQ further exposes differences in temporal
interpretation. For two clips labeled
\(\mathrm{BG}\to\mathrm{TTM}\), MiniCPM-o 4.5 predicts
\(\mathrm{SELF}\to\mathrm{TTM}\) in both Core tasks.
As shown in Fig.~\ref{fig:qualitative_cases}, its description choices preserve the event order
in the kitchen clip but reverse it in the card-game clip.
{These findings show how the three evaluation components reveal complementary failures
in transition identification, directional detection,
and temporal ordering.}



\begin{figure}[t]
    \centering
    \includegraphics[width=0.6\linewidth]{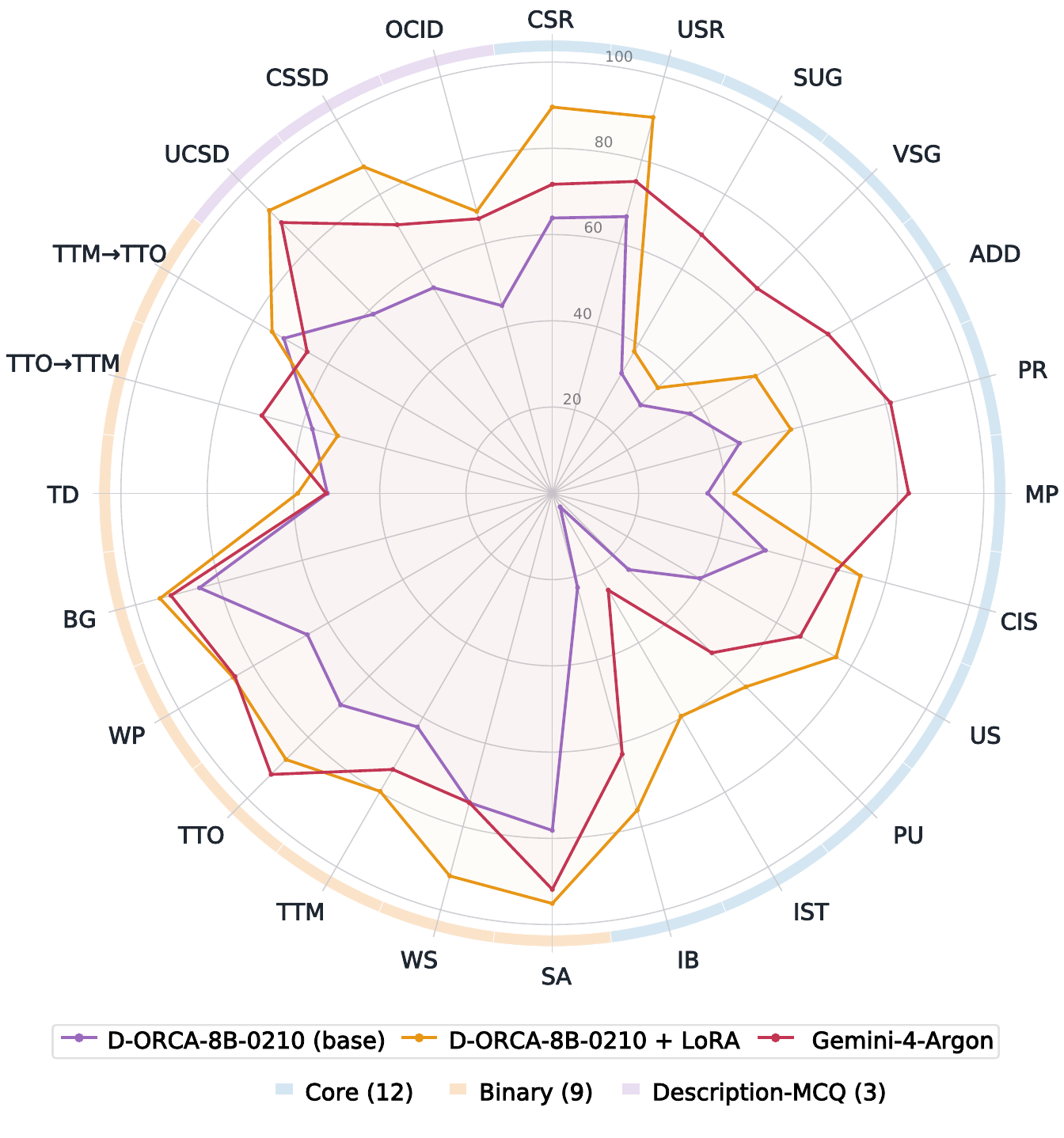}
    \vspace{-3mm}
    \caption{{Task-level accuracy before and after LoRA
adaptation of D-ORCA-8B-0210.}
    Gemini-4-Argon is a strong reference.}
    \label{fig:dorca_lora_subtasks}
    \vspace{-7mm}
\end{figure}

\subsection{Post-training for Social Understanding}
\label{sec:post_training}

We further investigate whether supervised post-training on the MSIU-Bench training split improves multimodal social interaction
understanding. We adapt four open-weight MLLMs from
different model families and sizes using
LoRA~\cite{hu2022lora} and compare
their performance before and after adaptation on the
same held-out test set.

\vspace{1mm}
\noindent
\noindent\textbf{Post-training enables small open-weight MLLMs
to approach frontier-model performance.}
Supervised adaptation improves overall Core accuracy
by 25.36--30.81 percentage points across four models.
Notably, Qwen2.5-Omni-3B improves from 24.36\% to
55.11\%, surpassing all open-weight direct-inference
baselines. Gemma-4-12B-it achieves the largest gain,
rising from 28.38\% to 59.19\%.
More remarkably, the adapted D-ORCA-8B-0210 achieves
62.95\% Core accuracy, approaching the performance
of Gemini-4-Argon (66.35\%), a highly capable
proprietary frontier MLLM, with a gap of only
3.40 percentage points. It even surpasses
Gemini-4-Argon on Binary accuracy
(79.90\% vs.\ 77.39\%).
These results demonstrate that targeted post-training
can substantially narrow the performance gap between
relatively small open-weight models and frontier MLLMs
on social interaction understanding.

\vspace{1mm}
\noindent
\textbf{Task-level comparisons reveal gaps that persist after post-training.}
Despite substantial overall gains, improvements vary considerably
across tasks (Fig.~\ref{fig:dorca_lora_subtasks}). D-ORCA-8B-0210 improves from
39.45\% to 75.94\% on utterance-state association and
from 25.00\% to 63.46\% on tracking changes in wearer
participation. In contrast, speaker--utterance grounding
and visual speaker grounding improve by only 5.88 and
5.66 percentage points, respectively. Even after
adaptation, D-ORCA trails Gemini-4-Argon by more than
30 percentage points on both tasks, indicating persistent
difficulty in associating utterances with the correct
participants and their visual identities.
These results suggest that {while
post-training substantially improves interaction-state
reasoning, accurate speaker grounding remains a key
bottleneck in multimodal social interaction understanding.}

These results demonstrate the potential of MSIU-Bench
for improving multimodal social interaction understanding.

\section{Conclusion}
MSIU-Bench provides a systematic framework for evaluating multimodal social interaction understanding, from participant grounding and conversational relationships to dynamic interaction-state reasoning. Our evaluation reveals persistent challenges in grounding speech and visual evidence in participant identities and tracking how social interactions evolve over time, even for frontier MLLMs. While post-training substantially improves overall performance, accurate speaker grounding remains challenging. By publicly releasing the benchmark and training data, we aim to advance research on robust multimodal grounding and dynamic social reasoning, supporting the development of more socially capable MLLMs.

\section*{LLM Usage Statement}

We used OpenAI's ChatGPT to assist with drafting and
revising the manuscript, including language refinement,
and to support the preparation of code and figures.
For dataset construction, Qwen3-VL-2B-Instruct was
used to generate visual descriptions, and
Qwen2.5-Omni-7B was used to assist with description
rewriting, as described in Section~3.4. The models used
for evaluation and post-training are specified in
Section~4. The authors reviewed the resulting content
and take responsibility for the final submission.

{
    \small
    \bibliographystyle{ieeenat_fullname}
    \bibliography{main}
}


\clearpage
\appendix
\setcounter{figure}{0}
\setcounter{table}{0}
\setcounter{equation}{0}
\renewcommand{\thefigure}{S\arabic{figure}}
\renewcommand{\thetable}{S\arabic{table}}
\renewcommand{\theequation}{S\arabic{equation}}
\renewcommand{\theHfigure}{supp.\arabic{figure}}
\renewcommand{\theHtable}{supp.\arabic{table}}
\renewcommand{\theHequation}{supp.\arabic{equation}}
\raggedbottom
\setcounter{topnumber}{3}
\setcounter{dbltopnumber}{2}
\renewcommand{\topfraction}{0.95}
\renewcommand{\dbltopfraction}{0.95}
\renewcommand{\textfraction}{0.05}
\renewcommand{\floatpagefraction}{0.60}
\renewcommand{\dblfloatpagefraction}{0.60}
\setlength{\floatsep}{10pt plus 2pt minus 2pt}
\setlength{\dblfloatsep}{10pt plus 2pt minus 2pt}
\setlength{\textfloatsep}{12pt plus 2pt minus 2pt}
\setlength{\dbltextfloatsep}{12pt plus 2pt minus 2pt}
\makeatletter
\setlength{\@fptop}{0pt}
\setlength{\@fpbot}{0pt plus 1fil}
\setlength{\@fpsep}{12pt}
\setlength{\@dblfptop}{0pt}
\setlength{\@dblfpbot}{0pt plus 1fil}
\setlength{\@dblfpsep}{12pt}
\makeatother

\twocolumn[{%
  \begin{center}
    {\Large\bfseries Supplementary Material\par}
  \end{center}
  \vspace{0.5em}
}]
\section{Detailed Task Definitions}
\label{sec:supp_task_definitions}

MSIU-Bench comprises three capability levels and three
evaluation components, with 12 \textbf{Core} tasks,
9 \textbf{Binary} tasks, and 3 \textbf{Description-MCQ}
tasks (description-based multiple-choice questions).
This section describes their prediction targets,
temporal anchors, and answer categories, organized
by sub-capability. A sub-capability may correspond
to multiple Core or Binary tasks.
Tasks are assigned to levels according to their
primary focus, although answering a question may
require capabilities from more than one level.
Representative task examples are provided in
Section~\ref{sec:supp_question_examples}.

For wearer-centered tasks, we use four interaction
states: $\mathrm{SELF}$ denotes wearer speech;
$\mathrm{TTM}$ denotes speech from another person
directed toward the wearer;
$\mathrm{TTO}$ denotes speech from another person
directed toward a non-wearer; and
$\mathrm{BG}$ denotes background conditions or
situations with no clear addressee.
For multiparty dialogue tasks, participant identifiers
refer to specific people in the current conversation.


\subsection{Level 1: Social Signal Perception and Identity Grounding}
\label{sec:supp_level1}

L1 assesses whether models can detect speaking
activity and associate speaker roles, marked
utterances, and people in the visual scene with
participant identities.

\Needspace{4\baselineskip}
\noindent
\textbf{P1. Speaking Activity and Speaker Detection.}
Two tasks determine whether anyone is speaking
at the end of the clip and whether the camera
wearer is speaking, respectively.
The current speaker-role task further distinguishes
\emph{Wearer}, \emph{Other}, and \emph{No speaker}.

\Needspace{4\baselineskip}
\noindent
\textbf{P2. Speaker--Utterance Grounding.}
These tasks attribute a marked utterance to its
speaker.
The wearer-centered task distinguishes whether
the utterance comes from the \emph{Wearer} or
\emph{Other}, while the multiparty dialogue task
identifies the speaker using participant identifiers.
The speaker of the marked utterance may differ
from the speaker at the end of the clip.

\Needspace{4\baselineskip}
\noindent
\textbf{P3. Visual Speaker Grounding.}
This task associates the speaker of a marked
utterance with the corresponding color-marked
person in the video.
The colors serve as participant identifiers
and do not denote clothing or appearance.


\subsection{Level 2: Current Social Interaction and Referential Reasoning}
\label{sec:supp_level2}

L2 assesses addressees, participant references,
wearer participation, and interaction states
at a specified moment or marked utterance.

\Needspace{4\baselineskip}
\noindent
\textbf{G1. Addressee and Interaction Target Reasoning.}
In wearer-centered settings, two Binary tasks
determine whether another person is addressing
the camera wearer or a non-wearer at the end
of the clip, respectively.
In multiparty dialogue, models identify the
specific addressee of the marked utterance.
The addressee is the intended recipient of the
utterance. Since the same speaker can address
different people, identifying the speaker alone
is insufficient.

\Needspace{4\baselineskip}
\noindent
\textbf{G2. Pronoun Reference Resolution.}
Using the dialogue context and available
audio-visual evidence, models identify the
participant referred to by a pronoun such as
\emph{he}, \emph{she}, \emph{him}, or \emph{her}.
The referent may be neither the speaker nor
the addressee.

\Needspace{4\baselineskip}
\noindent
\textbf{G3. Mentioned Person Identification.}
This task identifies the participant whose name
is masked in an utterance.
Unlike pronoun resolution, this task recovers
a masked person mention.
The mentioned person may also differ from
the addressee.

\Needspace{4\baselineskip}
\noindent
\textbf{G4. Wearer Participation.}
This task determines whether the camera wearer
participates in the verbal interaction associated
with the marked utterance.
Under the benchmark's definition,
$\mathrm{SELF}$ and $\mathrm{TTM}$ indicate
participation, whereas $\mathrm{TTO}$ and
$\mathrm{BG}$ indicate non-participation.
The wearer therefore counts as a participant
when directly addressed, even without speaking.
This judgment concerns the local interaction
associated with the marked utterance.

\Needspace{4\baselineskip}
\noindent
\textbf{G5. Current Interaction-State Inference.}
Two tasks identify the interaction state from
$\mathrm{SELF}$, $\mathrm{TTM}$, $\mathrm{TTO}$,
and $\mathrm{BG}$, using either the end of the
clip or a marked utterance as the temporal anchor.
The former assesses the interaction at the clip
endpoint, while the latter associates a specific
utterance with its interaction state.
A separate background Binary task determines
whether the clip ends in $\mathrm{BG}$.


\subsection{Level 3: Dynamic Interaction-State Reasoning}
\label{sec:supp_level3}

L3 compares social information at different
temporal positions to assess changes in interaction
states, wearer participation, and speaking targets.
The tasks distinguish whether a change occurs,
its direction, and its type.

\Needspace{4\baselineskip}
\noindent
\textbf{S1. Wearer Participation Update.}
Following the participation definition in G4,
this task compares the wearer's participation
at earlier and later temporal positions.
It distinguishes entering participation,
leaving participation, participating at both
positions, and participating at neither position.
An interaction-state change does not necessarily
change participation.
For example,
$\mathrm{TTM}\rightarrow\mathrm{SELF}$
changes the wearer's role from addressee to
speaker, while the wearer remains a participant
at both positions.

\Needspace{4\baselineskip}
\noindent
\textbf{S2. Interaction-State Transition Reasoning.}
These tasks determine whether an interaction-state
transition occurs and identify its earlier and
later states and direction.
The Core task includes the 12 directed transitions
between distinct interaction states and a
\emph{No Transition} class, giving 13 classes
in total.
The corresponding Binary task determines whether
an interaction-state change occurs.
For example,
$\mathrm{SELF}\rightarrow\mathrm{TTM}$ and
$\mathrm{TTM}\rightarrow\mathrm{SELF}$
involve the same states but represent changes
in opposite directions.

\Needspace{4\baselineskip}
\noindent
\textbf{S3. Interaction Boundary Understanding.}
This task determines whether the interaction
in the clip starts, ends, or does neither,
with answer categories \emph{Started},
\emph{Ended}, and \emph{Neither}, respectively.

\Needspace{4\baselineskip}
\noindent
\textbf{S4. Directed Changes in Speaking Targets.}
Two Binary tasks separately assess whether
$\mathrm{TTO}\rightarrow\mathrm{TTM}$ or
$\mathrm{TTM}\rightarrow\mathrm{TTO}$ occurs.
The former indicates a shift in the addressee
from a non-wearer to the wearer, while the
latter indicates the reverse.
Both states involve non-wearer speech, so models
must compare the earlier and later addressees
to determine whether the specified directional
shift occurs.


\subsection{Description-MCQ: Joint Understanding of Social Facts}
\label{sec:supp_description_mcq}

The three Description-MCQ tasks combine related
social facts into natural-language descriptions.
Models select, from four options, the description
that best matches the video and the provided context.

\Needspace{4\baselineskip}
\noindent
\textbf{Utterance-Centered Social Description.}
This task jointly assesses speaker identity,
utterance content, and the participant relations
associated with a marked utterance.
Depending on the question, the relevant person
may be the addressee, a pronoun referent, or
a mentioned participant.
The task is assigned to L1 while also involving
the participant relations assessed at L2.

\Needspace{4\baselineskip}
\noindent
\textbf{Current Social-State Description.}
This task jointly assesses the visible scene
and the wearer-centered interaction state
at the end of the clip.
Models select a description that matches both
the scene information and the speaking relation.
The interaction states follow the definitions
of $\mathrm{SELF}$, $\mathrm{TTM}$,
$\mathrm{TTO}$, and $\mathrm{BG}$.
This task primarily corresponds to L2.

\Needspace{4\baselineskip}
\noindent
\textbf{Opening--Closing Interaction Description.}
This task compares the social situations at the
beginning and end of a clip.
Models determine whether a candidate description
correctly represents the interaction states
at these two positions and their temporal order.
For example, for a
$\mathrm{TTO}\rightarrow\mathrm{TTM}$
sample, the correct description indicates that
another person addresses a non-wearer at the
beginning and that another person addresses
the wearer at the end.
This task primarily corresponds to L3.
It assesses only the two endpoint states and
their temporal order, without requiring
reconstruction of intervening events.


\section{Dataset Construction and Human Review}
\label{sec:supp_construction}

This section supplements the construction procedure
described in the main paper, covering annotation
alignment, representative task examples, description
options, and quality control.

\subsection{Annotation Alignment and QA Construction}
\label{sec:supp_annotation_alignment}

For wearer-centered tasks, we align the temporal
state annotations from Online TTM~\cite{du2026onlinettm}
with clip endpoints, marked utterances, and annotated
transition events to derive task-specific answers.
For multiparty dialogue tasks, we use the speaker
identities, participant relations, and visual identity
markers associated with Online-MMSI~\cite{li2026onlinemmsi}
to construct questions about speakers, addressees,
and referenced participants.

Each question has a defined prediction target and
temporal anchor. A video clip may support multiple
tasks, but each answer is derived from the annotations
corresponding to that question's target.
In particular, a marked utterance and the clip
endpoint may correspond to different interaction
states. We retain source and temporal metadata
to support verification against the original evidence.

\subsection{Representative Task Examples}
\label{sec:supp_question_examples}

We use multiple natural-language phrasings for each
task while retaining its prediction target and answer
space. Questions are accompanied by task-specific
context when needed.
Figures~\ref{fig:supp_l1_tasks}--\ref{fig:supp_description_tasks}
organize the 24 task illustrations into four groups:
six Core and Binary tasks in L1, nine in L2, six in L3,
and three Description-MCQ tasks.
These groups cover 12 Core, 9 Binary, and 3 Description-MCQ tasks.
Each panel includes selected video frames, the question,
candidate answers, and, where applicable, the marked utterance.
The panels illustrate task inputs and candidate answers;
correct answers are not highlighted.
Frames are arranged in temporal order from left to right.

These illustrations show task formats and sample-specific
answer options. Participant and color identifiers are local
to each sample. Selected frames do not reproduce the full
audiovisual context.

\begin{TaskFigure}[!t]
    \TaskPanel{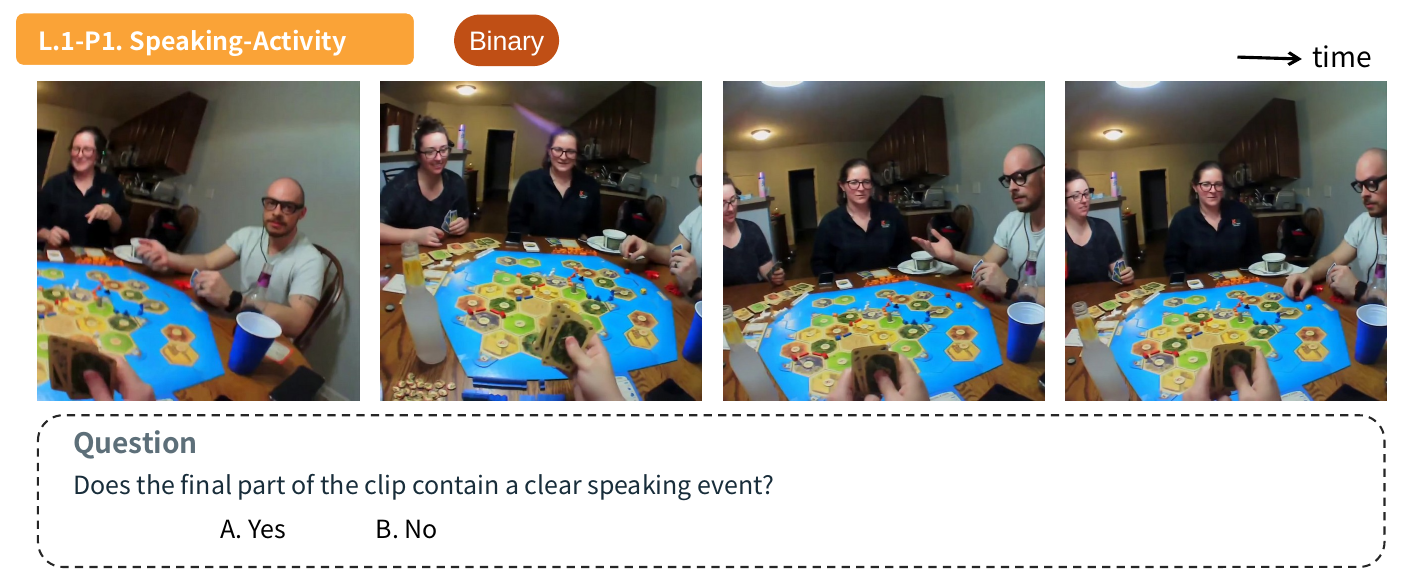}{fig:supp_speaking_activity}
    \TaskPanelGap
    \TaskPanel{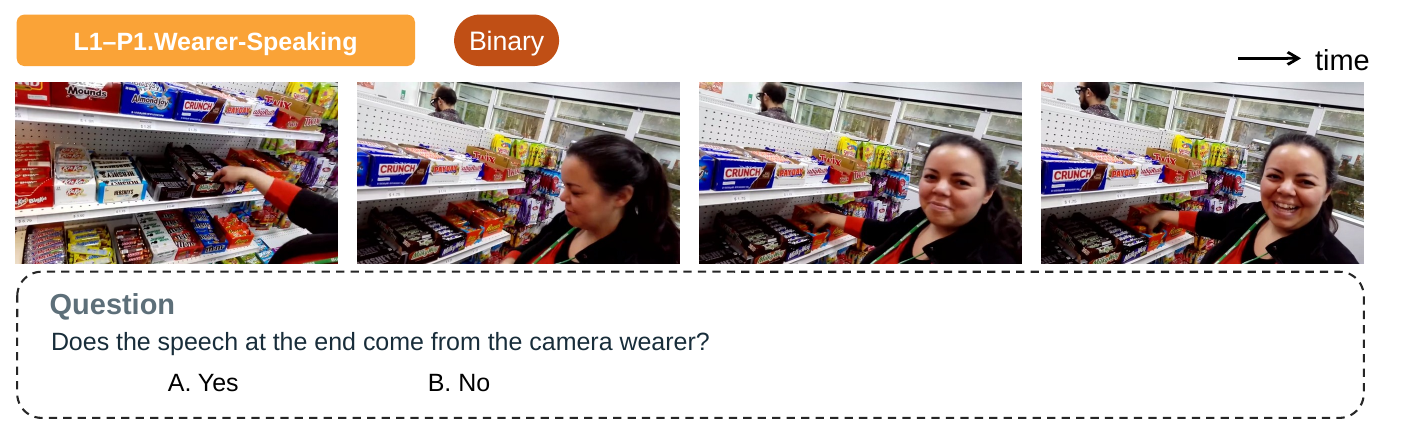}{fig:supp_wearer_speaking}
    \caption{Task illustrations for L1: social signal perception and identity grounding.}
    \label{fig:supp_l1_tasks}
\end{TaskFigure}

\subsection{Description Options and Distractors}
\label{sec:supp_description_construction}

Description-MCQ combines structured social annotations
with scene information from sampled video frames.
We use Qwen3-VL-2B-Instruct~\cite{bai2025qwen3vl}
to generate visual descriptions for the initial
candidates and Qwen2.5-Omni-7B~\cite{xu2025qwen25omni}
for subsequent description rewriting.
Source annotations provide the reference speaker
identities, participant relations, and interaction states.

Each question contains one correct description and
three distractors. Distractors may modify speaker
attribution, participant relations, scene details,
utterance content, or the order of endpoint states.
For the opening--closing task, descriptions concern
the states at the beginning and end of a clip and
their temporal order, without requiring reconstruction
of intervening events.
Figure~\ref{fig:supp_description_tasks} illustrates the candidate
descriptions for these three tasks.

\subsection{Automated Checks and Human Review}
\label{sec:supp_human_review}

Automated checks examine required fields, video paths,
and answer formats. Human review checks whether
the labeled answers agree with the available video,
audio, and textual context.

For Core and Binary tasks, reviewers check the
participant, temporal anchor, and social judgment
specified by each question against the corresponding
evidence. For Description-MCQ, reviewers examine all
four candidate descriptions to assess whether the
labeled answer is supported and uniquely correct.
They record errors, ambiguities, and suggested
corrections for subsequent checking.

\IfFileExists{fig/review_interface.pdf}{%
    Figure~\ref{fig:supp_review_interface} shows an example
    of the review interface.
}{}


\begin{TaskFigure}
    \ContinuedFloat
    \captionsetup{labelformat=continued,list=off}
    \TaskPanel{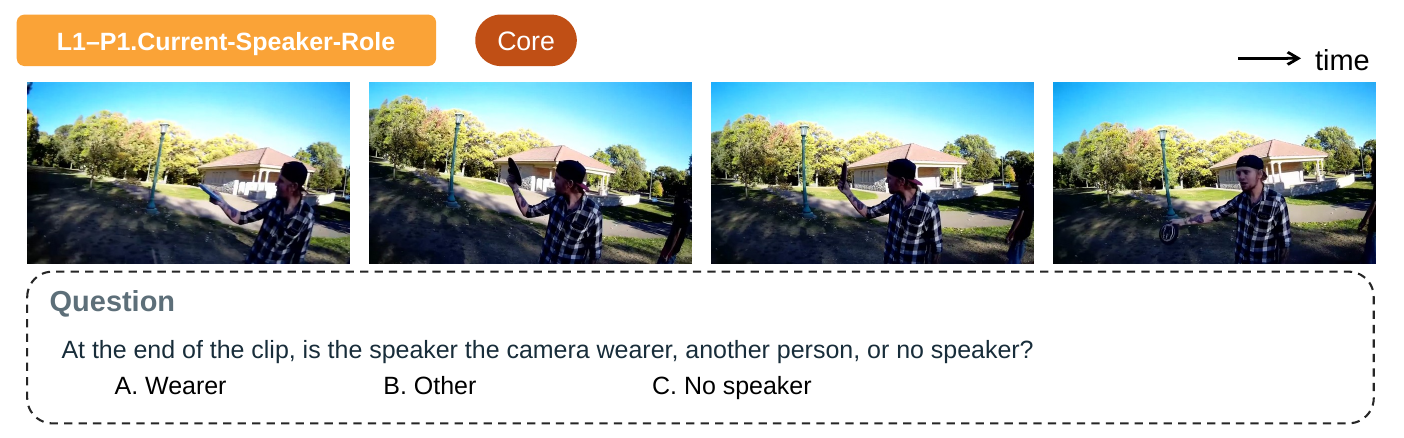}{fig:supp_current_speaker}
    \TaskPanelGap
    \TaskPanel{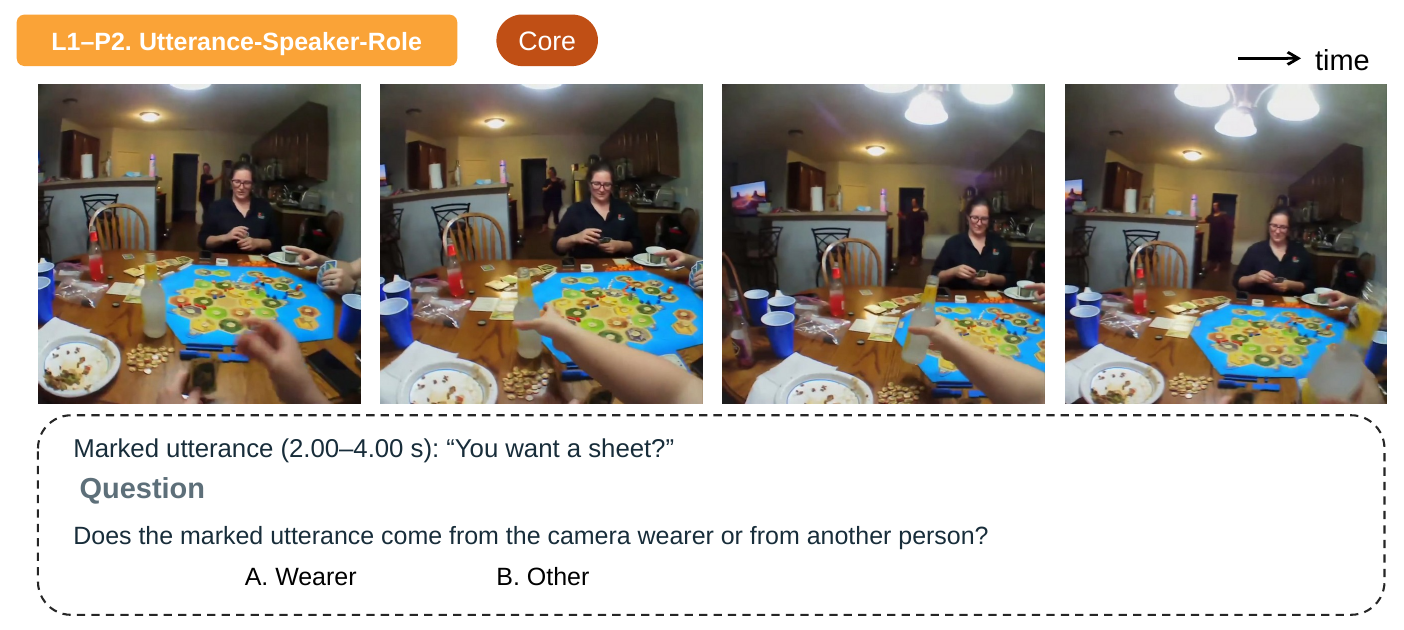}{fig:supp_utterance_speaker}
    \TaskPanelGap
    \TaskPanel{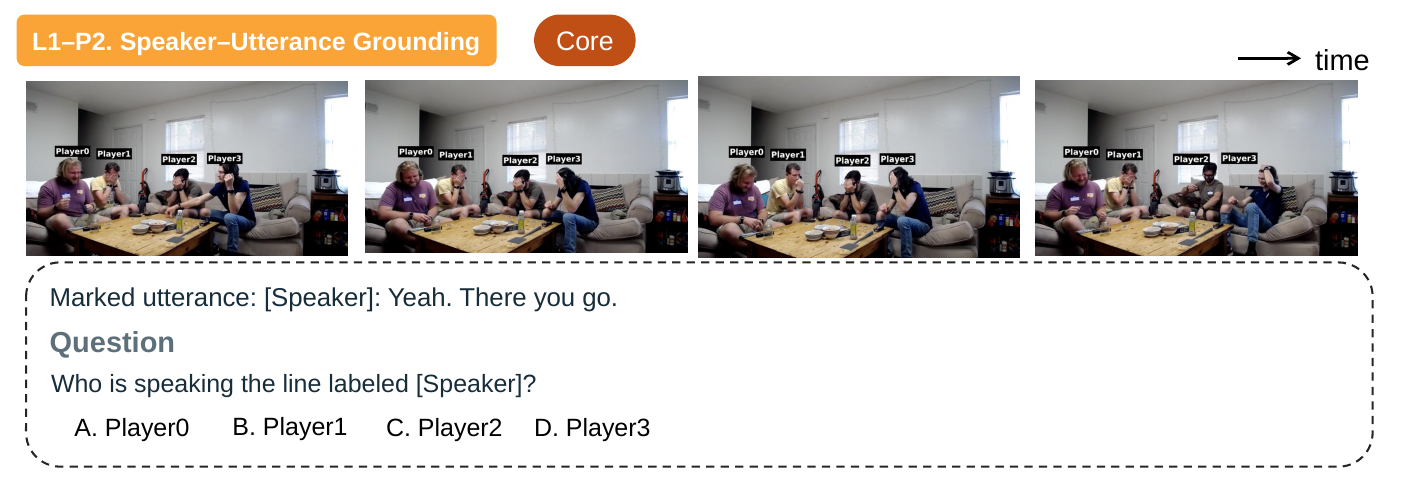}{fig:supp_speaker_utterance}
    \TaskPanelGap
    \TaskPanel{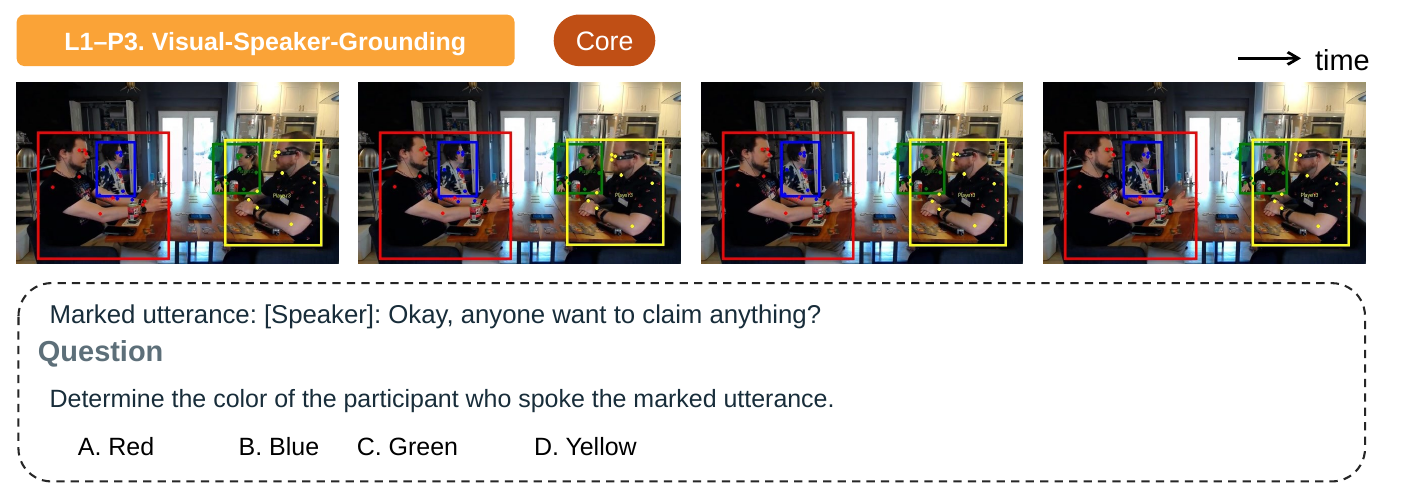}{fig:supp_visual_speaker}
    \caption{Task illustrations for L1: social signal perception and identity grounding.}
    \label{fig:supp_l1_tasks_continued_2}
\end{TaskFigure}

\begin{TaskFigure}
    \TaskPanel{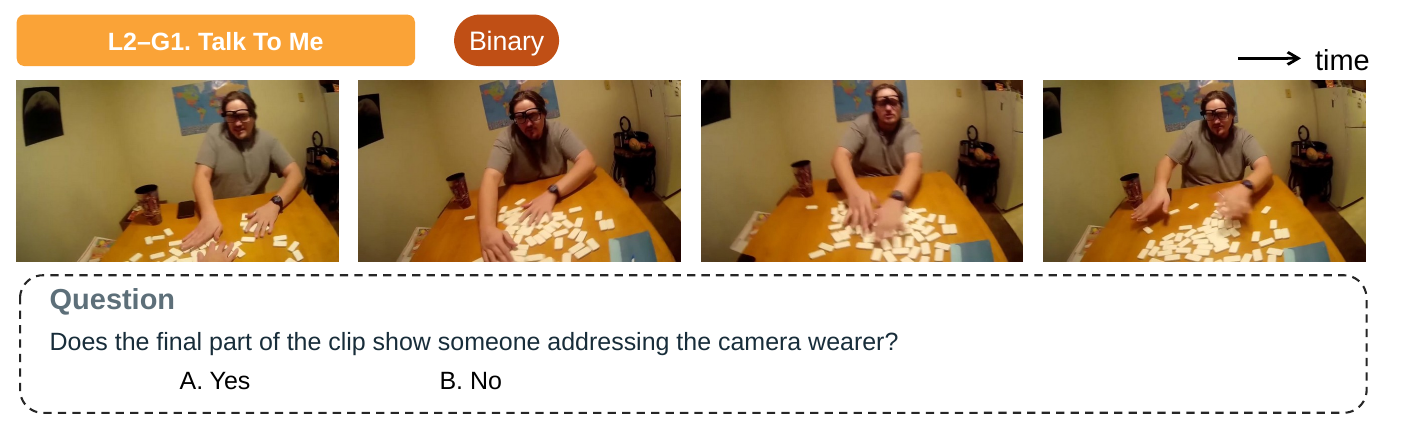}{fig:supp_ttm}
    \TaskPanelGap
    \TaskPanel{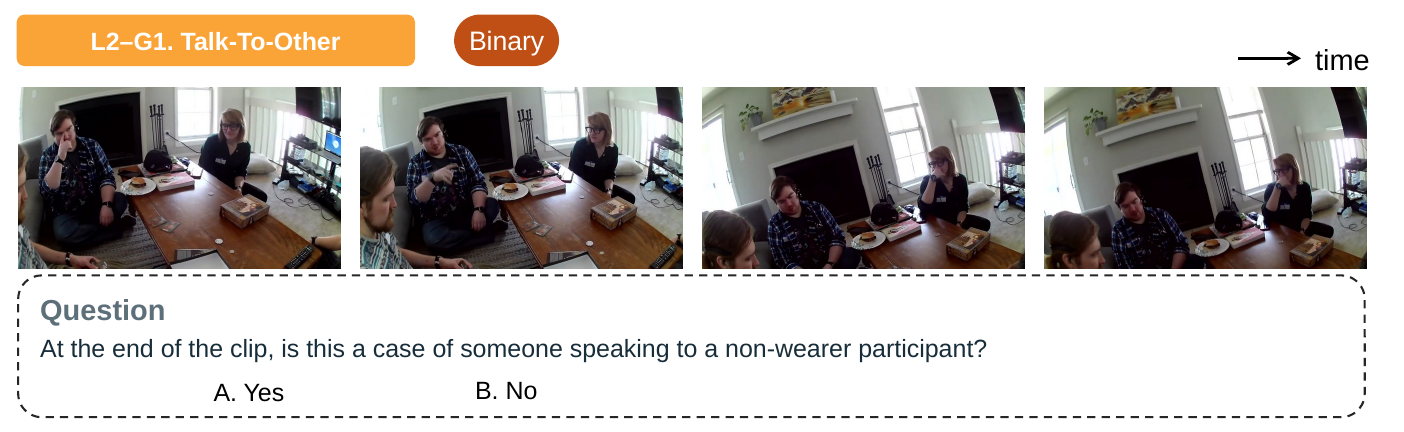}{fig:supp_tto}
    \TaskPanelGap
    \TaskPanel{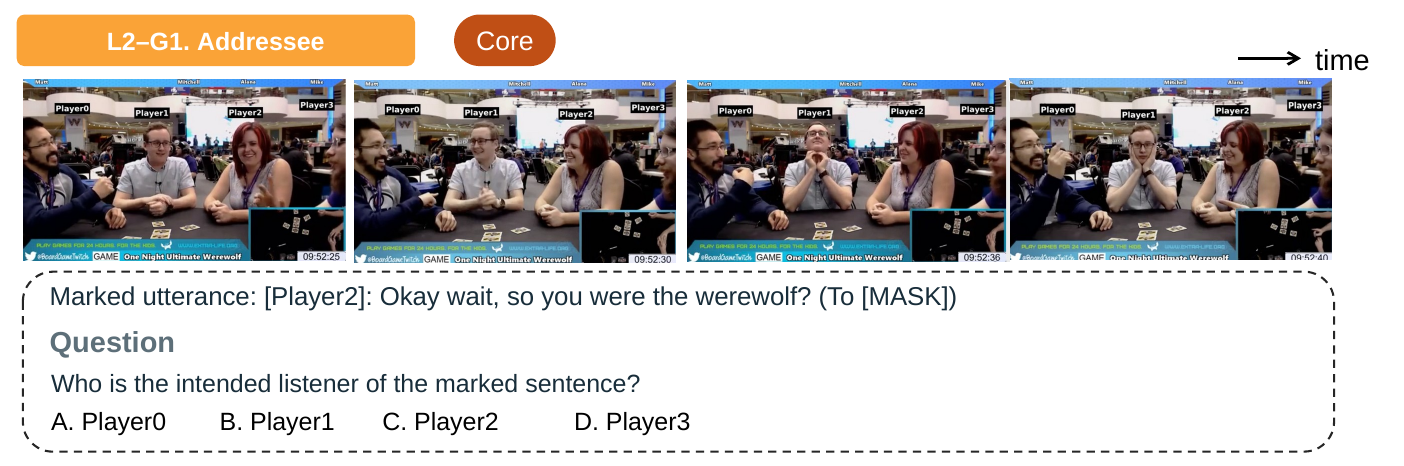}{fig:supp_addressee}
    \TaskPanelGap
    \TaskPanel{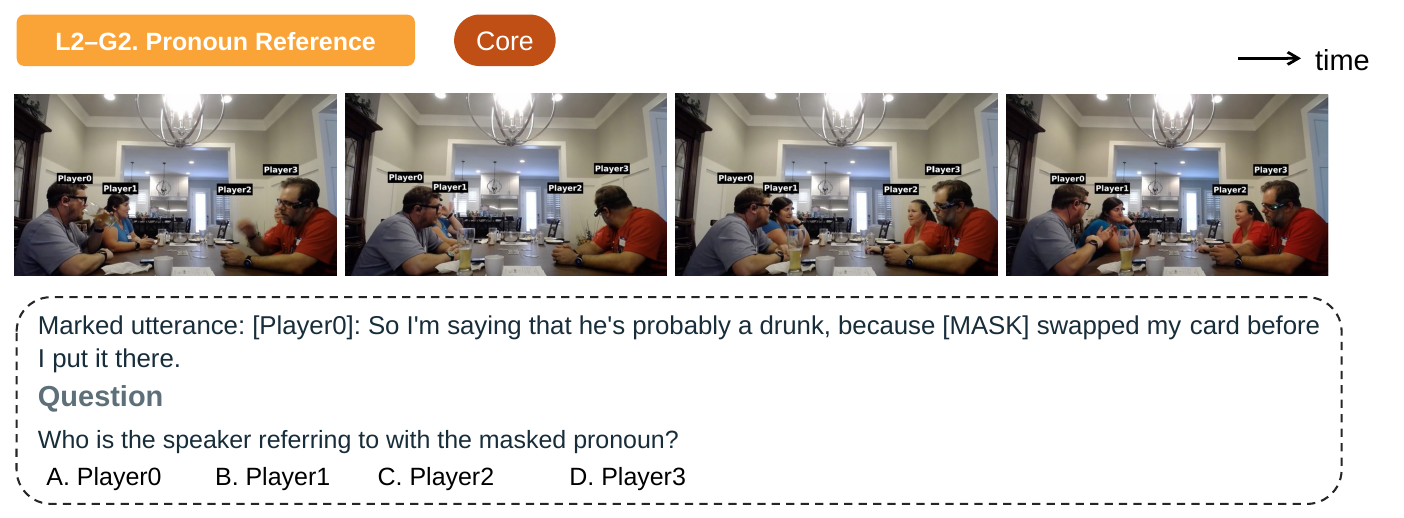}{fig:supp_pronoun}
    \caption{Task illustrations for L2: current social interaction and referential reasoning.}
    \label{fig:supp_l2_tasks}
\end{TaskFigure}

\begin{TaskFigure}
    \ContinuedFloat
    \captionsetup{labelformat=continued,list=off}
    \TaskPanel{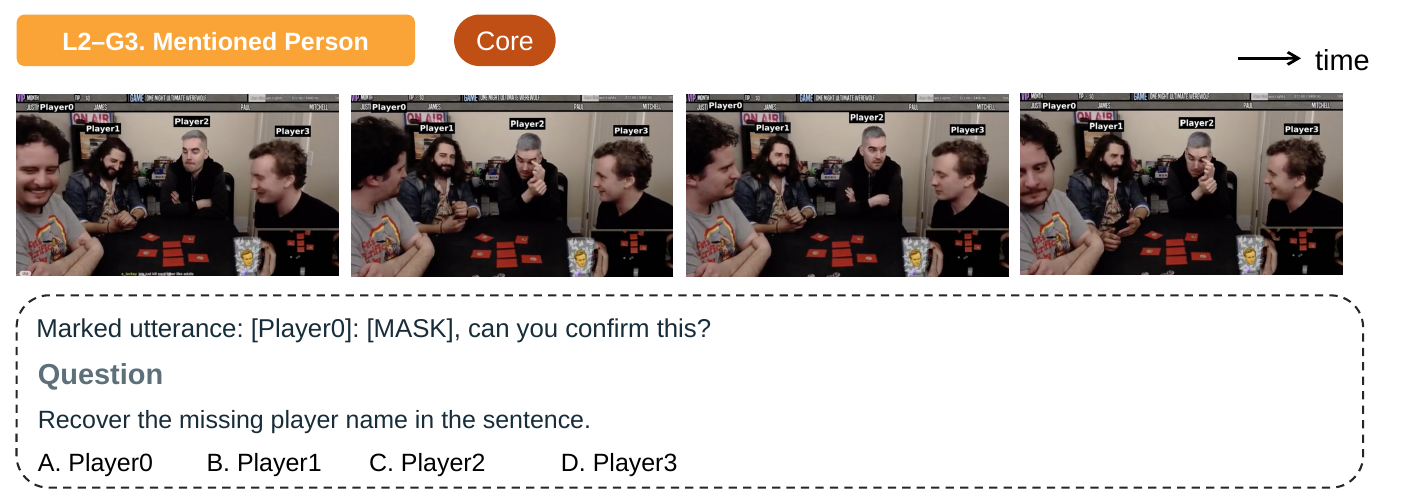}{fig:supp_mentioned_person}
    \TaskPanelGap
    \TaskPanel{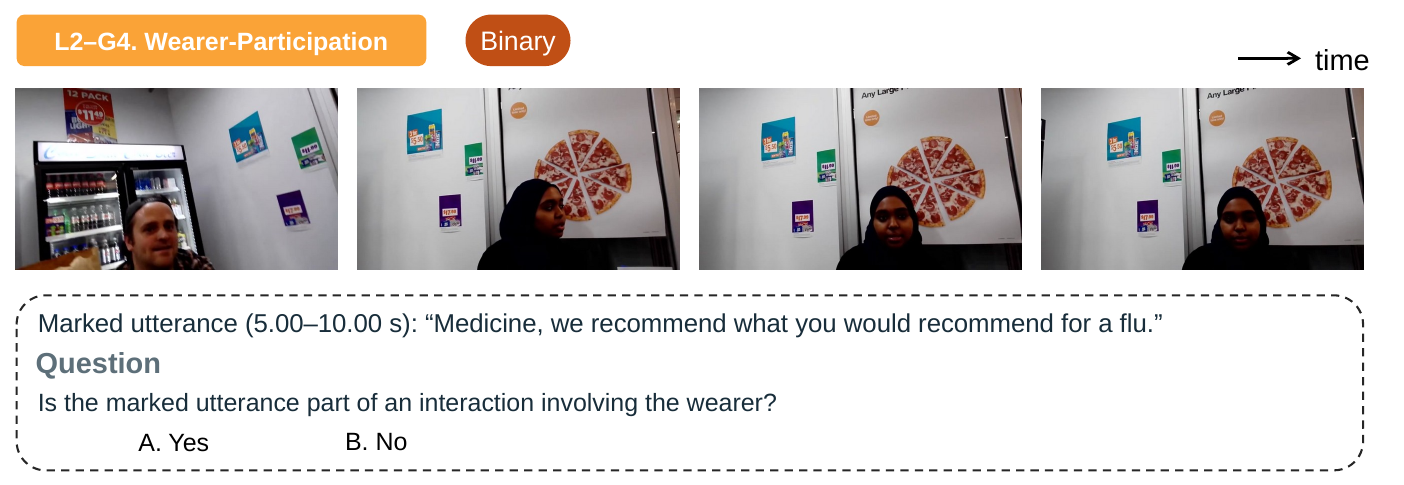}{fig:supp_wearer_participation}
    \TaskPanelGap
    \TaskPanel{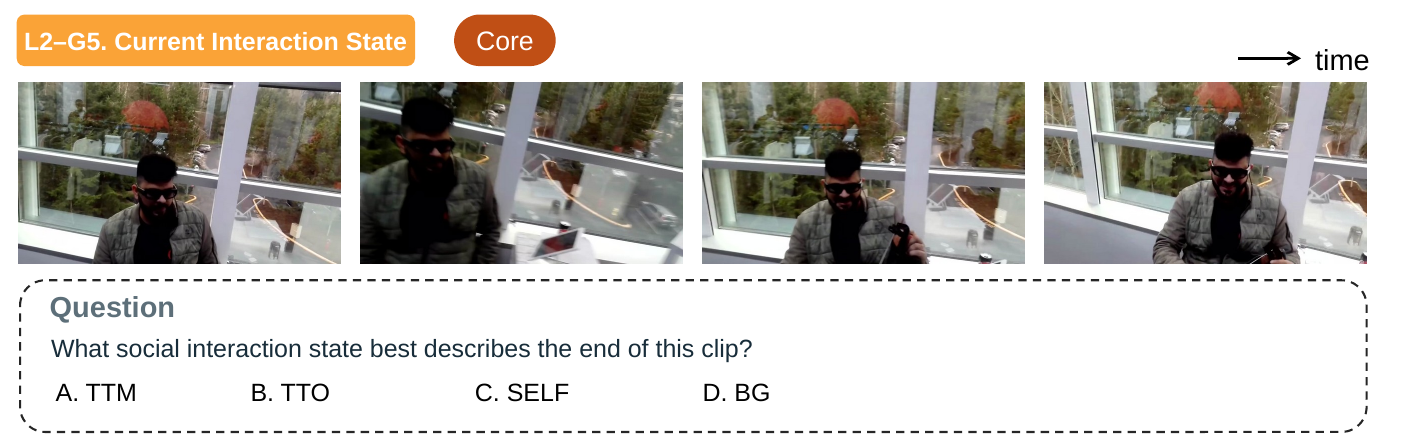}{fig:supp_current_state}
    \TaskPanelGap
    \TaskPanel{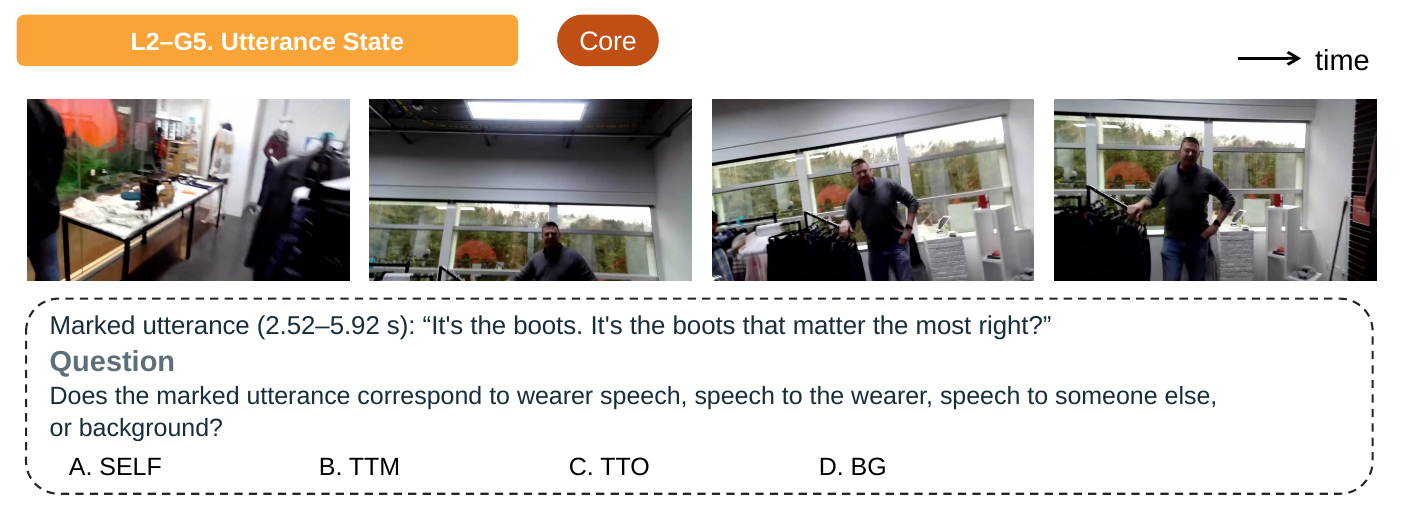}{fig:supp_utterance_state}
    \caption{Task illustrations for L2: current social interaction and referential reasoning.}
    \label{fig:supp_l2_tasks_continued_2}
\end{TaskFigure}

\begin{TaskFigure}
    \ContinuedFloat
    \captionsetup{labelformat=continued,list=off}
    \TaskPanel{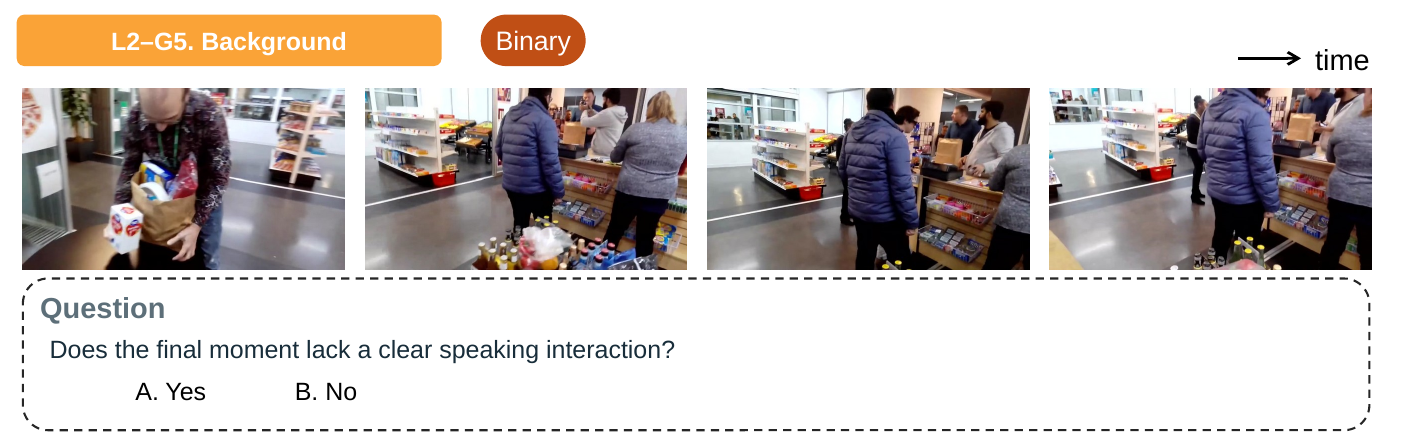}{fig:supp_background}
    \caption{Task illustrations for L2: current social interaction and referential reasoning.}
    \label{fig:supp_l2_tasks_continued_3}
\end{TaskFigure}

\begin{TaskFigure}
    \TaskPanel{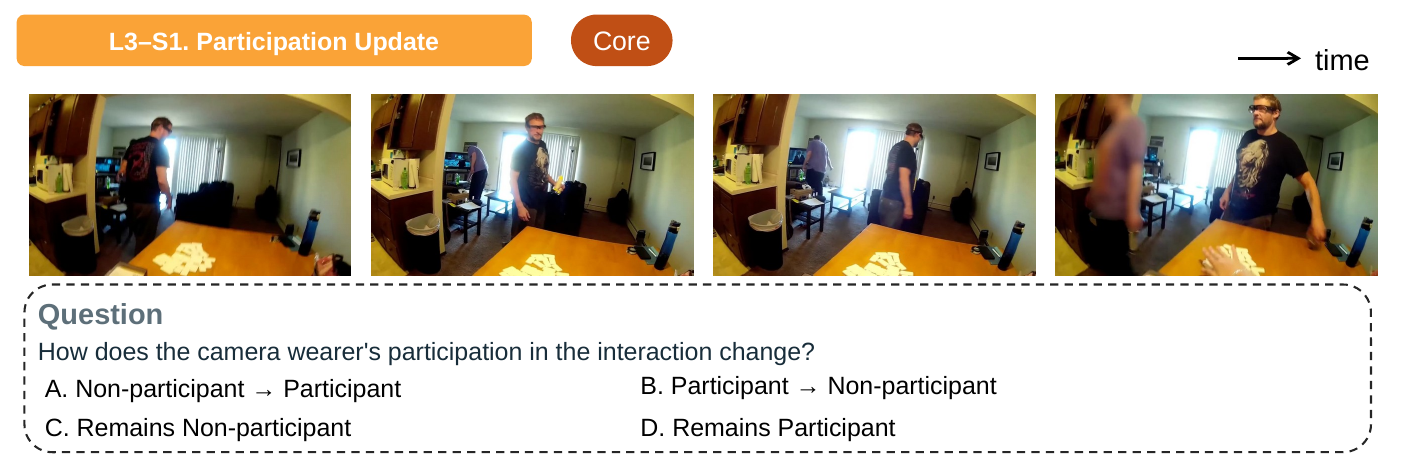}{fig:supp_participation_update}
    \TaskPanelGap
    \TaskPanel{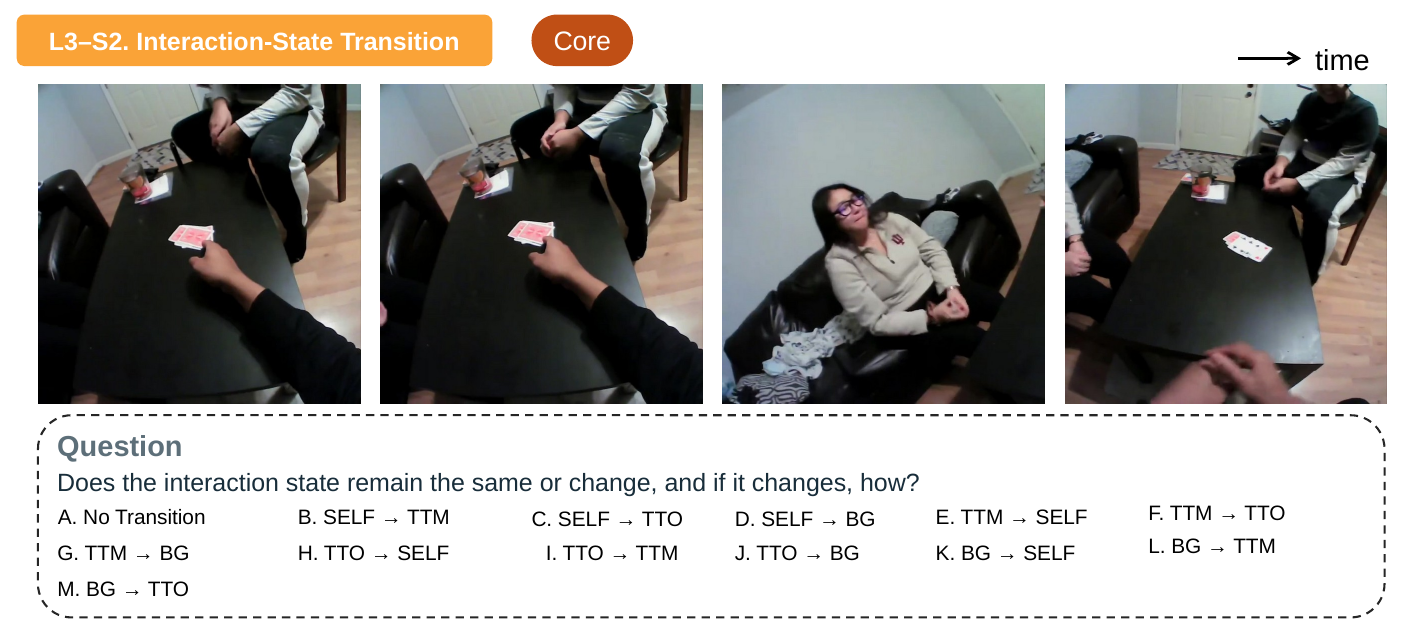}{fig:supp_state_transition}
    \TaskPanelGap
    \TaskPanel{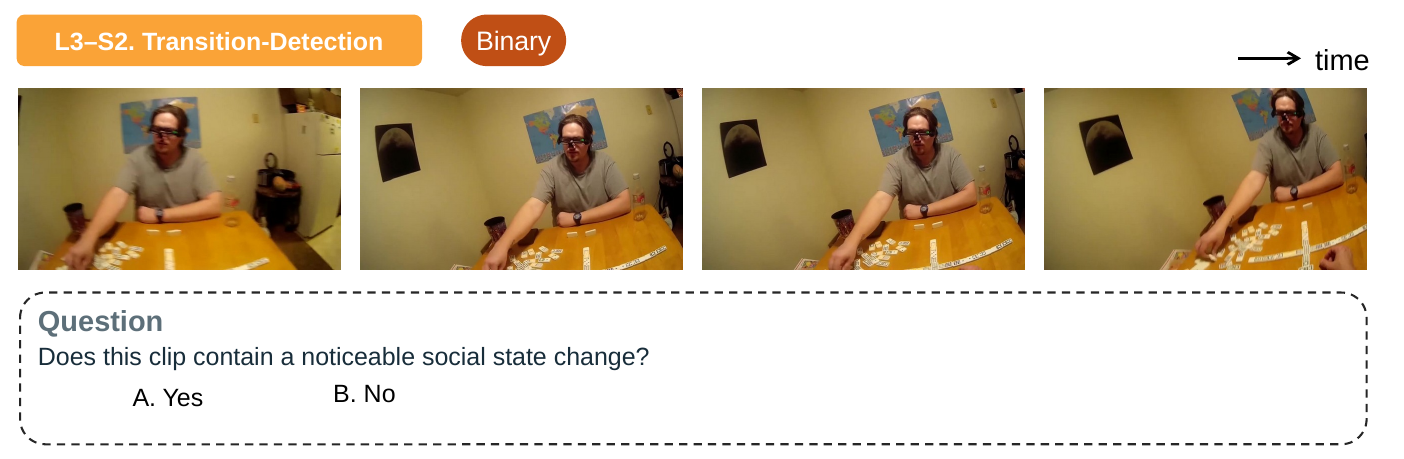}{fig:supp_transition_detection}
    \caption{Task illustrations for L3: dynamic interaction-state reasoning.}
    \label{fig:supp_l3_tasks}
\end{TaskFigure}

\begin{TaskFigure}
    \ContinuedFloat
    
    \captionsetup{labelformat=continued,list=off}
    \TaskPanel{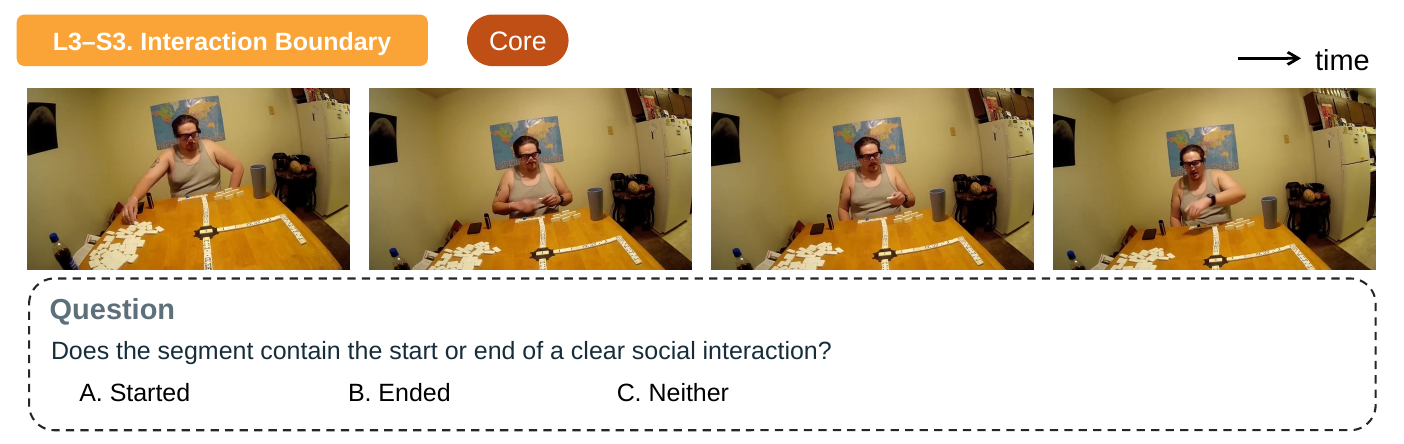}{fig:supp_interaction_boundary}
    \TaskPanelGap
    \TaskPanel{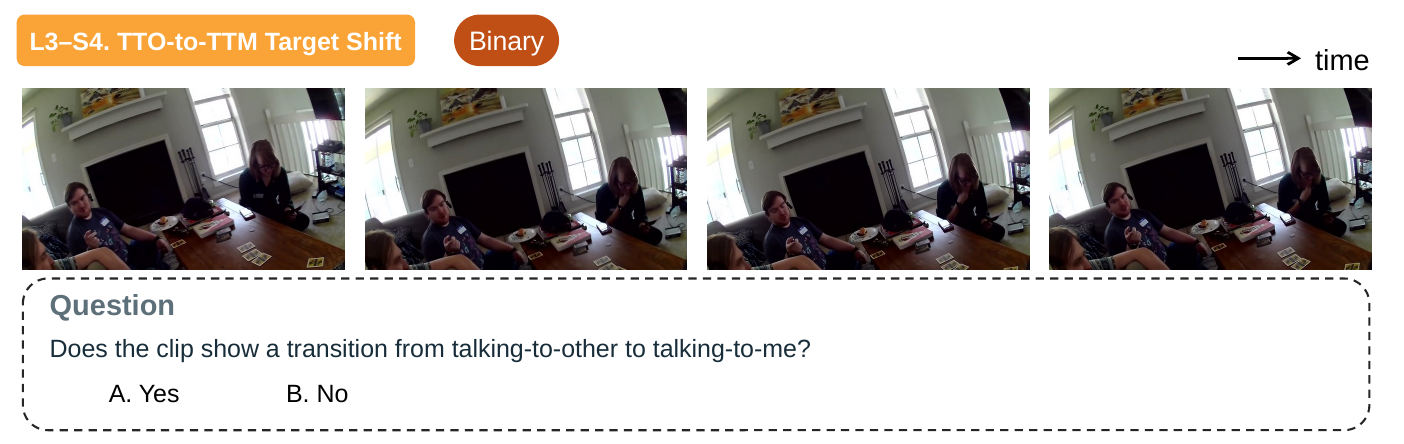}{fig:supp_tto_ttm}
    \TaskPanelGap
    \TaskPanel{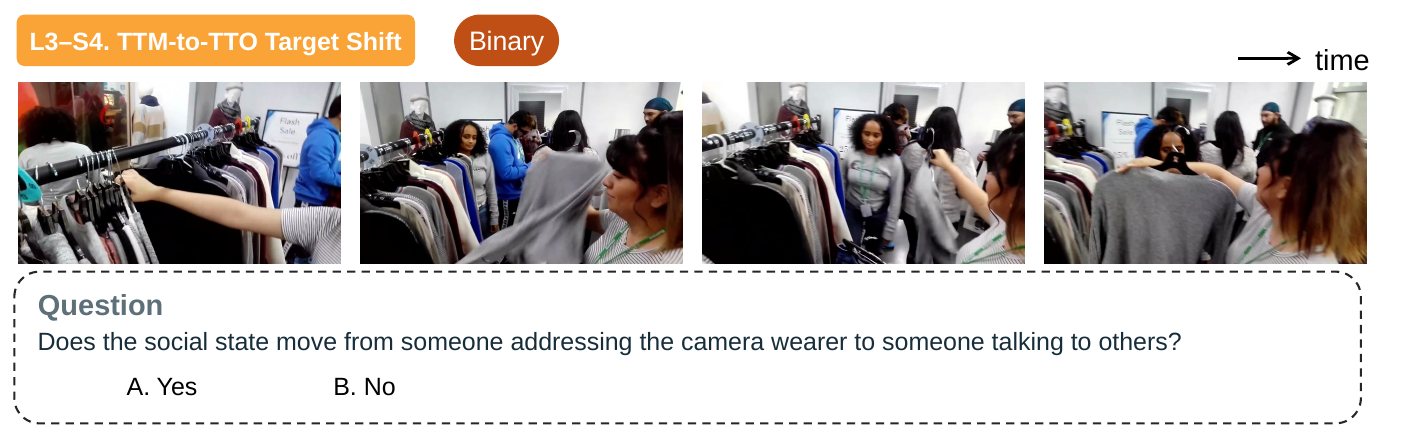}{fig:supp_ttm_tto}
    \caption{Task illustrations for L3: dynamic interaction-state reasoning.}
    \label{fig:supp_l3_tasks_continued_2}
\end{TaskFigure}

\begin{TaskFigure}
    \TaskPanel[\linewidth]{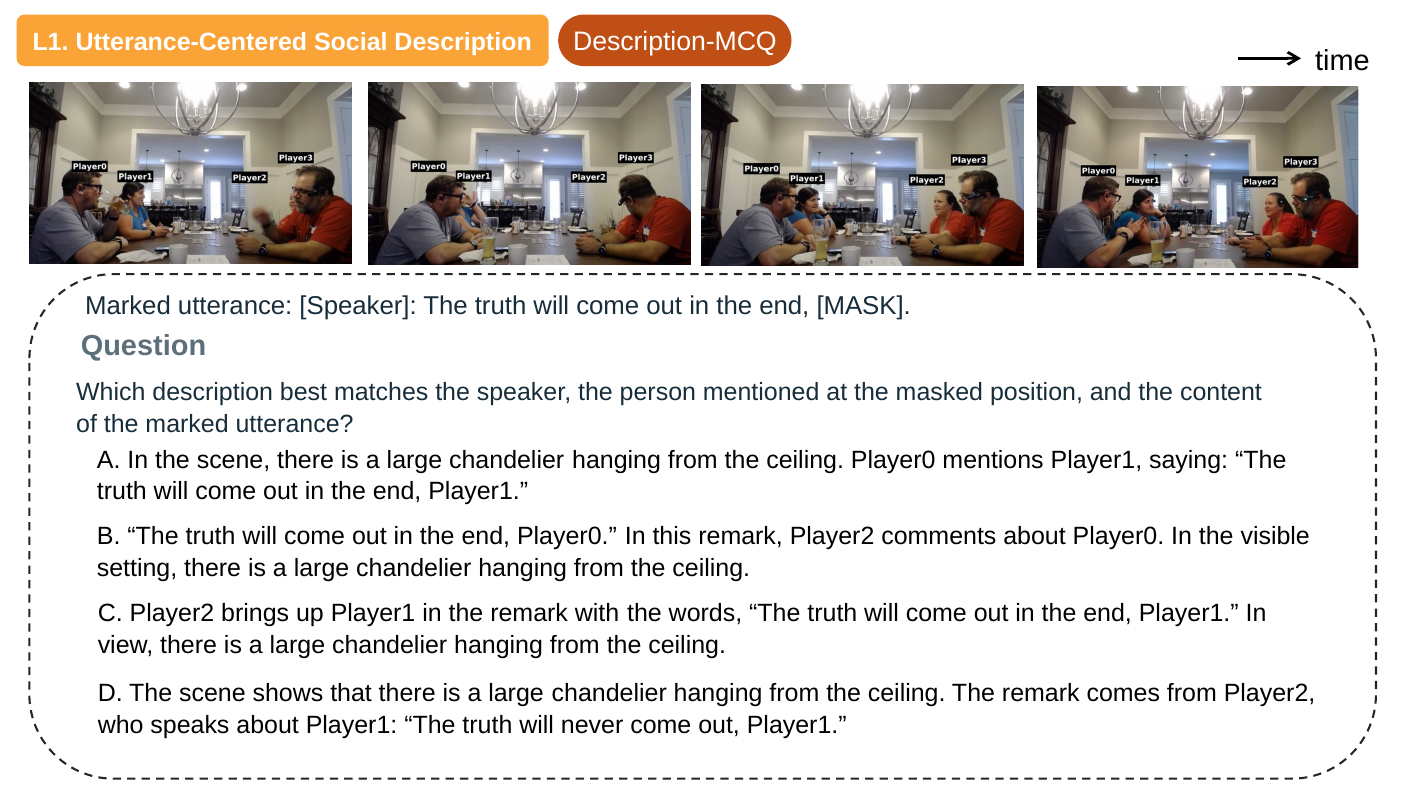}{fig:supp_utterance_description}
    \TaskPanelGap
    \TaskPanel[\linewidth]{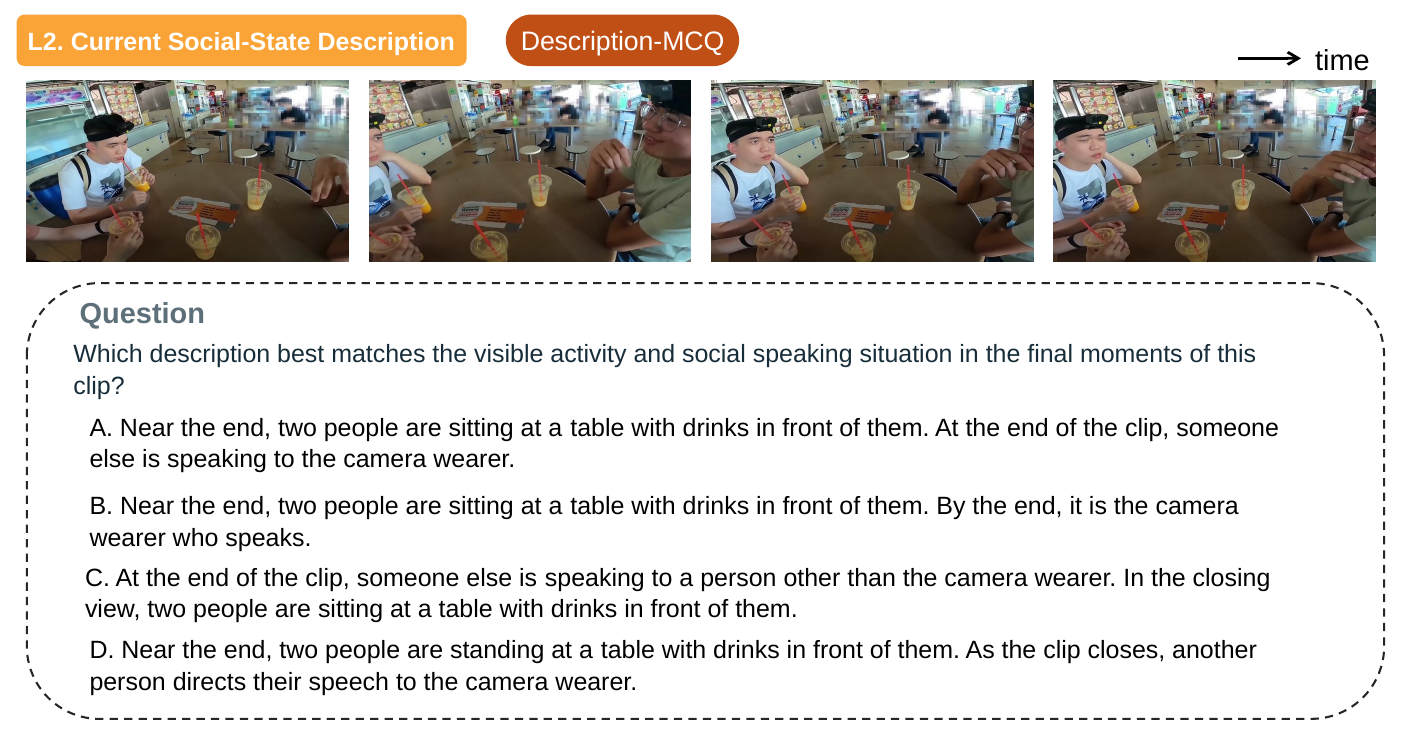}{fig:supp_current_description}
    \caption{Task illustrations for the three Description-MCQ tasks.}
    \label{fig:supp_description_tasks}
\end{TaskFigure}

\WideFigurePage{
\ContinuedFloat
    \captionsetup{labelformat=continued,list=off}
    \TaskPanel[\linewidth]{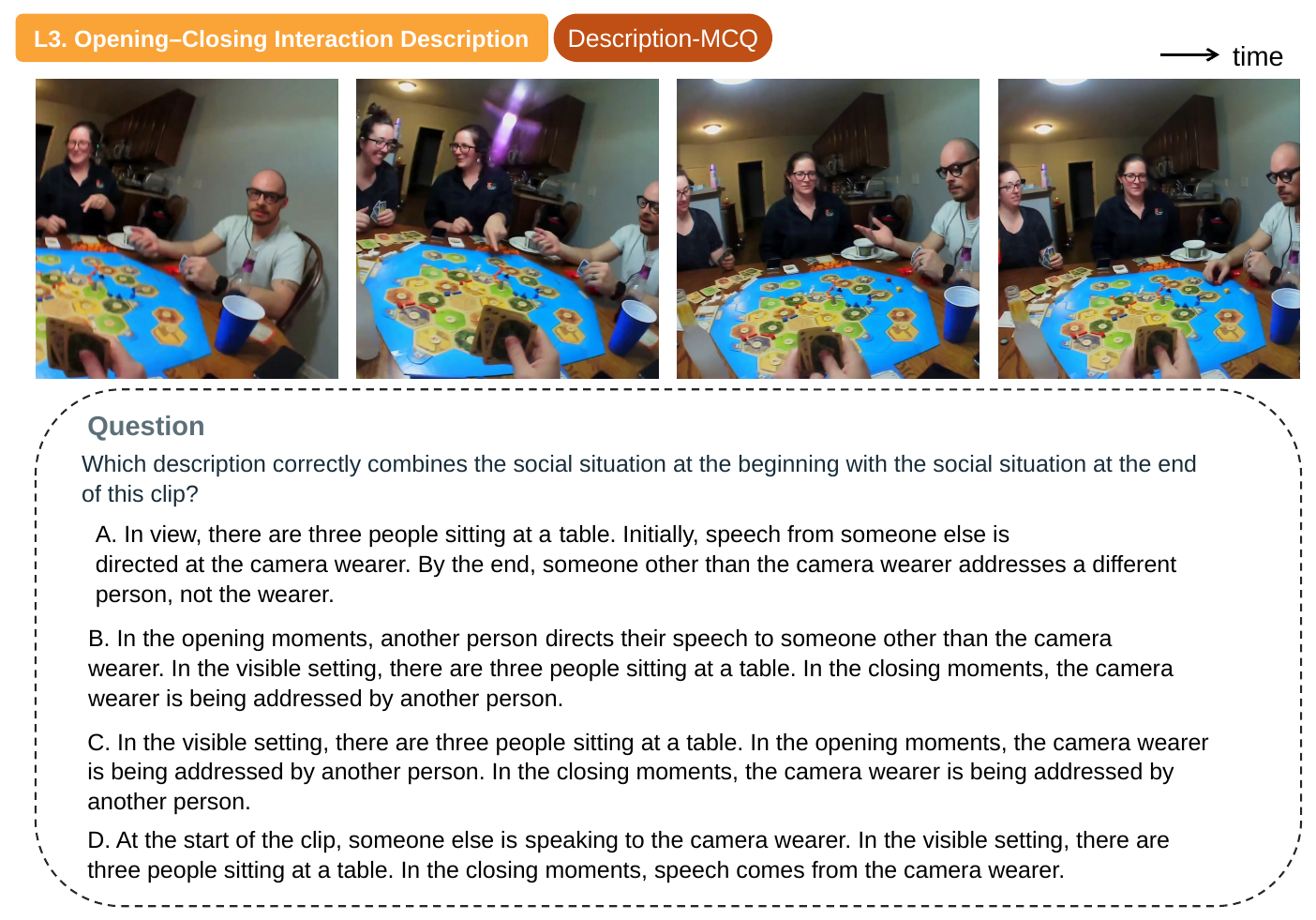}{fig:supp_opening_closing_description}
    \caption{Task illustrations for the three Description-MCQ tasks.}
    \label{fig:supp_description_tasks_continued_2}
}

\FloatBarrier
\IfFileExists{fig/review_interface.pdf}{%
    \begin{figure*}[!htbp]
        \centering
        \includegraphics[clip,
            width=\linewidth,
            height=0.75\textheight,
            keepaspectratio
        ]{fig/review_interface.pdf}
        \caption{\textbf{Human review interface.}
        Example of the interface used to review question--answer samples.}
        \label{fig:supp_review_interface}
    \end{figure*}
      
}{}

\section{Experimental and Fine-Tuning Details}
\label{sec:supp_experimental_details}

This section provides additional implementation details on model
inputs, decoding, and LoRA fine-tuning.

\subsection{Direct Inference Settings}
\label{sec:supp_direct_inference}

Each model receives the video, question, answer choices, and any
task-specific textual context, without in-context examples.
The answer choice order for each question is kept fixed across models.
Audio is included according to each model's evaluation configuration,
and audiovisual inputs are processed using the corresponding model
interfaces.

\paragraph{Video processing.}
Qwen2.5-Omni and Qwen3-Omni-Instruct sample videos at 2~fps,
with a maximum of 64 frames and a per-frame pixel limit of
100{,}352.
D-ORCA samples frames at 0.5-second intervals, with a maximum
of 128 frames and adaptive resizing.
HumanOmni and VideoLLaMA2.1-AV uniformly sample 32 and 8 frames,
respectively, with a processed resolution of $384 \times 384$;
Ola and OmniVinci use 64 frames.
For video-SALMONN, Ming-Lite-Omni, Gemma, and
Qwen3-Omni-Thinking, the number of sampled frames depends on
clip duration, with a maximum of 768 frames, and the spatial
pixel budget is adjusted according to the frame count.
Gemini-4-Argon receives the original videos through its API,
using 1~fps sampling and the low media resolution setting.

\paragraph{Audio and decoding.}
Local audio waveform preprocessing uses a sampling rate of
16~kHz.
HumanOmni uses up to 30 seconds of audio, while
VideoLLaMA2.1-AV samples eight audio segments of 2 seconds each.
Gemini-4-Argon receives videos with their original audio tracks.
Most models in non-reasoning mode use greedy decoding with
a maximum generation length of 64 tokens.
Qwen3-Omni-Thinking uses sampling with a temperature of 0.6,
top-$p$ of 0.95, and top-$k$ of 20, with a maximum generation
length of 32{,}768 tokens, including reasoning tokens.
Gemini-4-Argon uses the \texttt{medium} thinking setting with
the same generation budget of 32{,}768 tokens.

\subsection{LoRA Fine-Tuning Settings}
\label{sec:supp_lora_settings}

We fine-tune Qwen2.5-Omni-7B, Qwen2.5-Omni-3B,
D-ORCA-8B-0210, and Gemma-4-12B-it using LoRA on
question--answer pairs from the MSIU-Bench training split.
We compare performance before and after adaptation on the
same fixed test set.
The validation split is used for checkpoint selection.
Table~\ref{tab:supp_lora_config} summarizes the LoRA fine-tuning
configurations of the four models.
LoRA is applied to the attention projection layers and
feed-forward networks in the language decoder.


\begin{table}[htbp]
    \centering
    \caption{LoRA fine-tuning configuration.}
    \label{tab:supp_lora_config}
    \small
    \setlength{\tabcolsep}{4pt}
    \renewcommand{\arraystretch}{1.15}
    \begin{tabularx}{\linewidth}{@{}p{0.39\linewidth}X@{}}
        \toprule
        \textbf{Parameter} & \textbf{Setting} \\
        \midrule
        LoRA rank/ alpha/ dropout
            & 8 / 16 / 0.05 \\
        Optimizer
            & AdamW \\
        Learning rate
            & $5 \times 10^{-5}$ \\
        Weight decay
            & 0 \\
        Effective batch size
            & 8 \\
        Learning rate schedule
            & 3\% linear warmup, followed by cosine decay \\
        \bottomrule
    \end{tabularx}
\end{table}

\FloatBarrier
\section{Additional Results with Balanced Metrics}
\label{sec:supp_results}

This section supplements Sec.~4 by reporting balanced accuracy (BA) and
macro-F1 for representative models on the structured tasks
(Core and Binary).

\subsection{Evaluation Metrics}
\label{sec:supp_metrics}

For a task $t$, let $\mathcal{C}_t^+$ denote the classes with nonzero
ground-truth support in the test set. For each class $c$, $\mathrm{TP}_{t,c}$,
$\mathrm{FP}_{t,c}$, and $\mathrm{FN}_{t,c}$ denote true positives,
false positives, and false negatives, respectively. We compute
\begin{align}
\mathrm{BA}_t
 &= \frac{1}{|\mathcal{C}_t^+|}
    \sum_{c\in\mathcal{C}_t^+}
    \frac{\mathrm{TP}_{t,c}}{\mathrm{TP}_{t,c}+\mathrm{FN}_{t,c}}, \\
\mathrm{MacroF1}_t
 &= \frac{1}{|\mathcal{C}_t^+|}
    \sum_{c\in\mathcal{C}_t^+}
    \frac{2\mathrm{TP}_{t,c}}
    {2\mathrm{TP}_{t,c}+\mathrm{FP}_{t,c}+\mathrm{FN}_{t,c}}.
\end{align}
Classes without ground-truth examples are excluded from these averages.
Invalid or missing final answers count as incorrect and contribute a false
negative for the ground-truth class. Following the task-macro aggregation
used in the main paper, we compute each metric per task and average it
equally across the 12 Core tasks or the 9 Binary tasks. All results are
reported as percentages.

\subsection{Results for Representative Models}
\label{sec:supp_balanced_results}

Table~\ref{tab:supp_balanced_metrics} reports BA and macro-F1
for five representative direct-inference models discussed in
Sec.~4.2, complementing the accuracy results in the main paper.

\begin{table}[!htbp]
\centering
\caption{Task-macro balanced accuracy (BA) and macro-F1 (\%)
for representative direct-inference models on Core and Binary.
Bold indicates the best result among the listed models.}
\label{tab:supp_balanced_metrics}
\small
\setlength{\tabcolsep}{2.5pt}
\footnotesize
\begin{tabular}{@{}lcccc@{}}
\toprule
& \multicolumn{2}{c}{Core}
& \multicolumn{2}{c}{Binary} \\
\cmidrule(lr){2-3}\cmidrule(lr){4-5}
Model & BA & Macro-F1 & BA & Macro-F1 \\
\midrule
Gemini-4-Argon
& \textbf{65.57} & \textbf{59.40}
& \textbf{70.72} & \textbf{66.52} \\
video-SALMONN 2 Pro 32B
& 36.16 & 30.36 & 61.24 & 57.73 \\
video-SALMONN 2 Pro 4B
& 35.95 & 29.87 & 57.46 & 49.19 \\
D-ORCA-8B-0210
& 30.57 & 27.38 & 57.95 & 55.13 \\
MiniCPM-o 4.5
& 34.23 & 29.72 & 58.53 & 54.01 \\
\bottomrule
\end{tabular}
\end{table}

Gemini-4-Argon achieves the highest BA and macro-F1 among
the selected models on both components.
Within the video-SALMONN 2 Pro family, the 32B model exceeds
the 4B model by only 0.21 and 0.49 percentage points in
Core BA and macro-F1, respectively, while its advantages
are larger on Binary.

\WideFigurePage{
\centering
    \includegraphics[clip,width=0.94\linewidth]{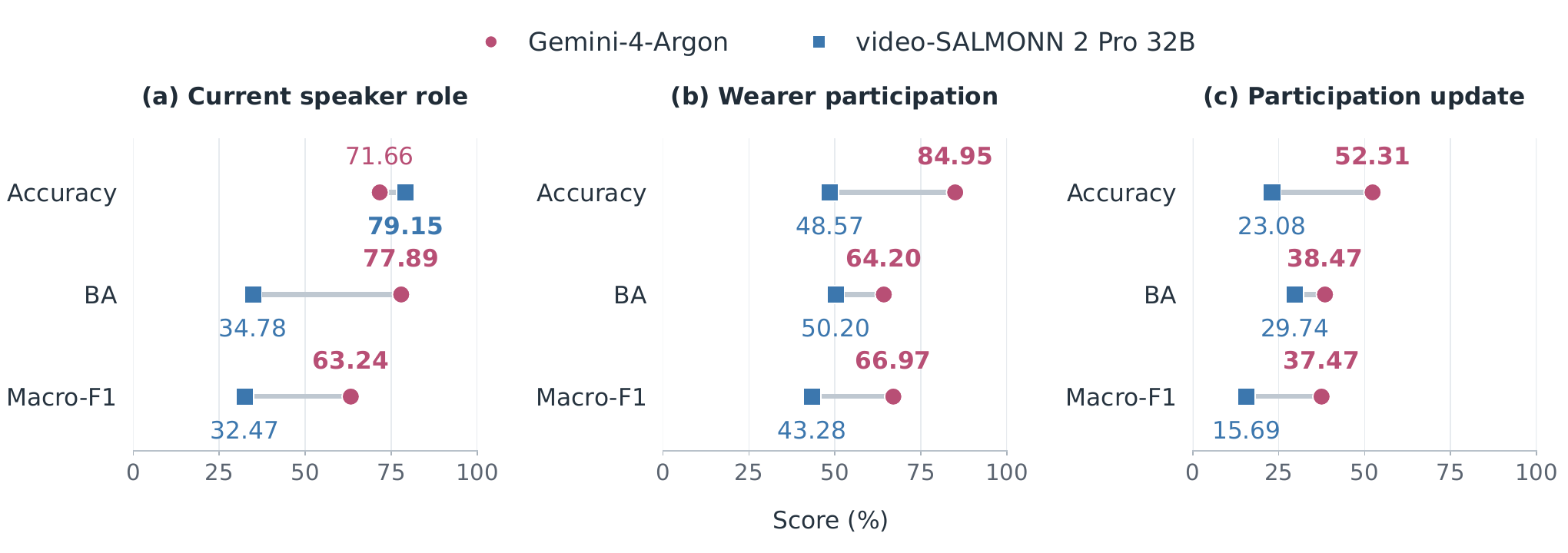}
    \caption{Comparison of accuracy, BA, and macro-F1 (\%) between
    Gemini-4-Argon and video-SALMONN 2 Pro 32B on
    (a) current speaker-role identification,
    (b) wearer participation, and
    (c) wearer participation update.
    Gray lines connect the two model scores for the same task and
    metric. Bold values indicate the higher score within each pair.}
    \label{fig:supp_task_metrics}
}

\begin{figure*}[!b]
    \centering
    \includegraphics[clip,width=0.64\textwidth]{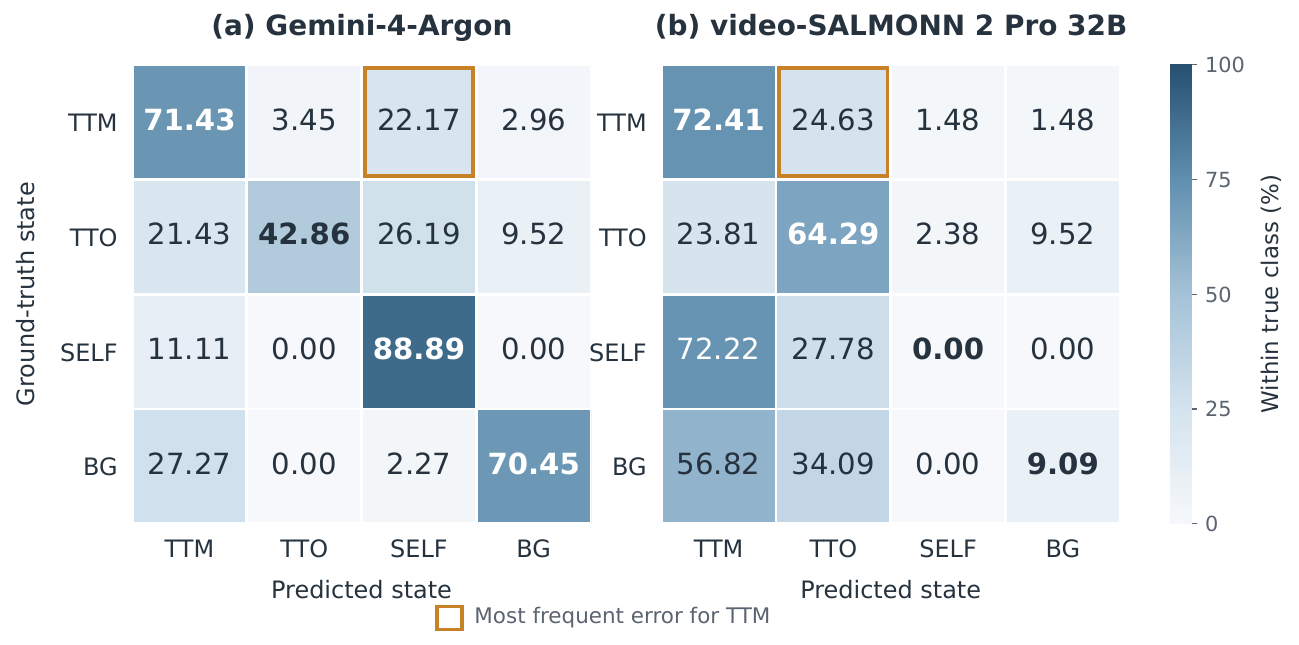}
    \caption{Row-normalized confusion matrices (\%) for
    Gemini-4-Argon and video-SALMONN 2 Pro 32B on current
    interaction-state classification. Rows indicate ground-truth
    states and columns indicate predicted states. Both panels
    share the same color scale. Orange outlines highlight the
    most frequent misclassification among ground-truth TTM
    examples for each model.}
    \label{fig:supp_state_confusion}
\end{figure*}

\subsection{Analysis of Selected Tasks}
\label{sec:supp_selected_task_analysis}

We compare Gemini-4-Argon and video-SALMONN 2 Pro 32B on three
tasks emphasized in Sec.~4.2. The latter is the open-weight
direct-inference baseline with the highest overall Core accuracy.
Figure~\ref{fig:supp_task_metrics} presents their accuracy, BA,
and macro-F1 on current speaker-role identification, wearer
participation, and wearer participation update.

On current speaker-role identification, Pro 32B achieves
higher accuracy than Argon but substantially lower BA
and macro-F1.
This ranking reversal supports the observation in Sec.~4.2
that higher accuracy can conceal uneven performance
across speaker roles.

Argon outperforms Pro 32B under all three metrics on
wearer participation and participation update.
However, its BA on wearer participation is 64.20\%,
compared with an accuracy of 84.95\%, consistent with
the asymmetric errors discussed in the main paper.
On participation update, its BA and macro-F1 reach
38.47\% and 37.47\%, respectively, indicating that
tracking changes in participation remains challenging.

\subsection{Confusion Analysis of Current Interaction States}
\label{sec:supp_state_confusion}

Figure~\ref{fig:supp_state_confusion} compares Gemini-4-Argon
and video-SALMONN 2 Pro 32B on four-class interaction-state
classification at the end of a clip, supplementing the analysis
in Sec.~4.2.

Among the 203 ground-truth TTM examples, Argon and Pro 32B
correctly classify 145 and 147, respectively. However, their
most frequent errors differ: Argon misclassifies 45 examples
as SELF, whereas Pro 32B misclassifies 50 as TTO. These results
show that similar TTM recognition rates can correspond to
different confusion patterns, involving the distinction
between speaker roles and between addressees, respectively.


\FloatBarrier
\WideFigurePage{
\centering
    \includegraphics[clip,width=\linewidth]{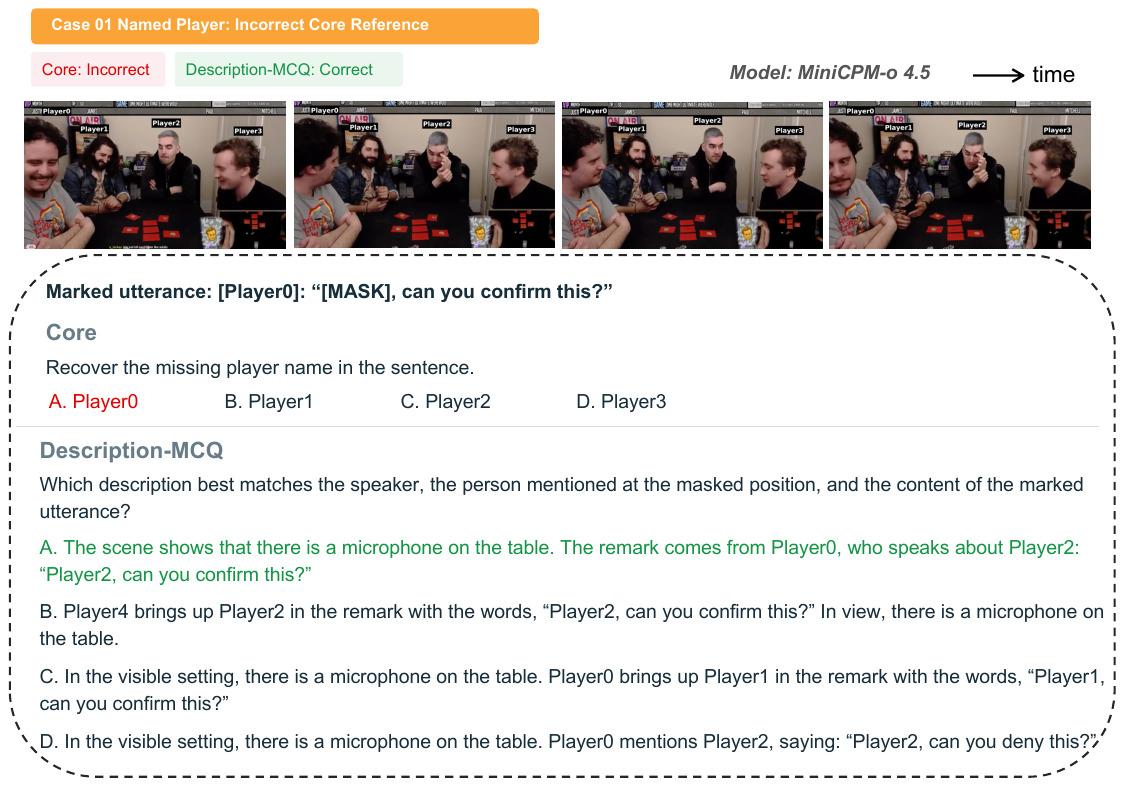}
    \caption{Person-reference example for MiniCPM-o 4.5.
    Core incorrectly identifies Player0 as the mentioned
    participant, whereas the selected Description-MCQ option
    correctly identifies Player2.
    Green and red indicate correct and incorrect
    model-selected answers, respectively.}
    \label{fig:supp_reference_case}
}

\subsection{Complementarity of Evaluation Components}
\label{sec:supp_component_complementarity}

We supplement Sec.~4.2 with three qualitative examples
illustrating the complementary diagnostic value of Core,
Binary, and Description-MCQ. All examples use MiniCPM-o 4.5.
Within each example, the related questions concern the same
marked utterance or the same social event in a clip.

\paragraph{Person reference.}
Figure~\ref{fig:supp_reference_case} compares Core and
Description-MCQ on a person-reference example. For the marked
utterance ``[MASK], can you confirm this?'', the speaker is
Player0 and the mentioned participant is Player2.
In Core, the model incorrectly fills the masked position with
Player0.
\newpage
\noindent
In Description-MCQ, however, it selects the correct
description, which associates the utterance with speaker
Player0 and mentioned participant Player2. The selected
description therefore contains the correct participant
information, while the direct person-reference answer is
incorrect. This error would be missed if only the description
choice were examined.


\WideFigurePage{
\centering
    \includegraphics[clip,width=\linewidth]
    {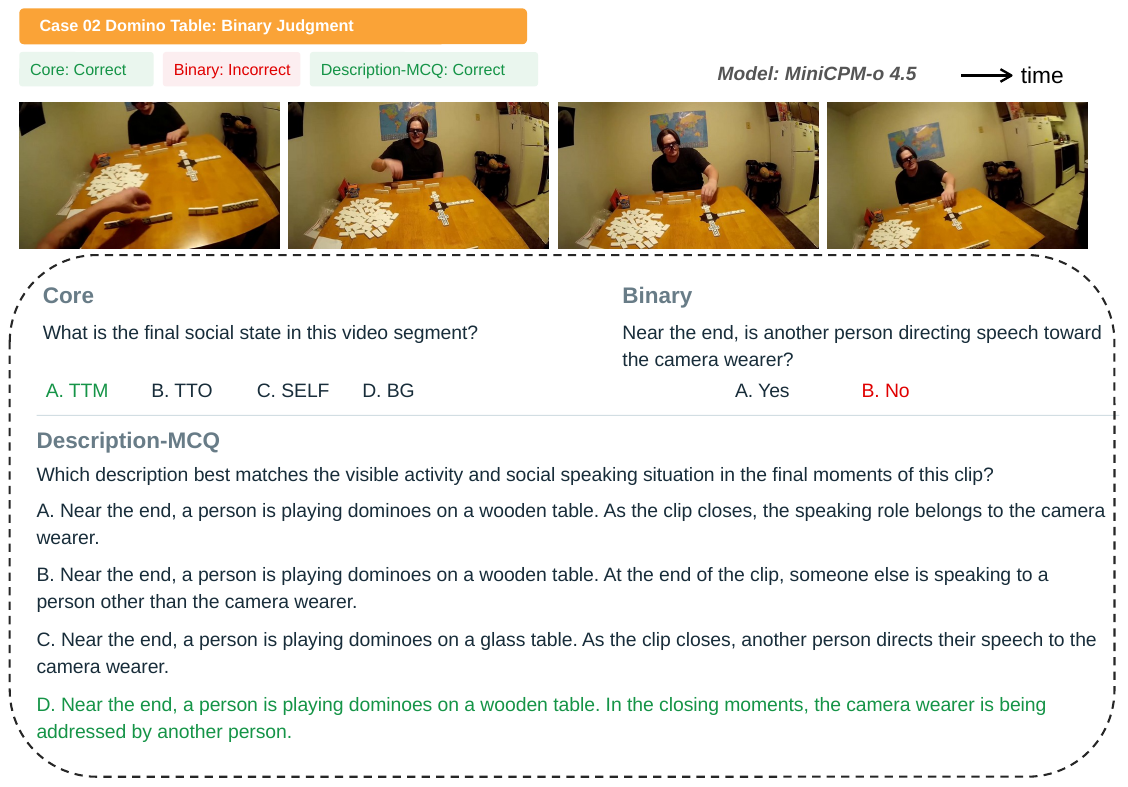}
    \caption{Current interaction-state example for MiniCPM-o 4.5.
    Core and Description-MCQ correctly identify speech directed
    to the camera wearer, but Binary incorrectly answers
    ``No'' to the corresponding question.
    Green and red indicate correct and incorrect
    model-selected answers, respectively.}
    \label{fig:supp_state_case}
}

\paragraph{Current interaction state.}
Figure~\ref{fig:supp_state_case} presents an example from
a domino-table clip. At the end of the clip, another
participant is speaking to the camera wearer. The model
correctly selects TTM in Core and chooses a Description-MCQ
option that captures this interaction together with the
wooden table and dominoes.
\newpage
\noindent
However, it answers ``No'' to the
corresponding Binary question asking whether another person
is addressing the wearer. Correct state classification and
description selection thus coexist with an incorrect judgment
of the directly related binary proposition. Binary provides
an additional diagnostic view of the otherwise correct
Core prediction.

\WideFigurePage{
\centering
    \includegraphics[clip,width=\linewidth]
    {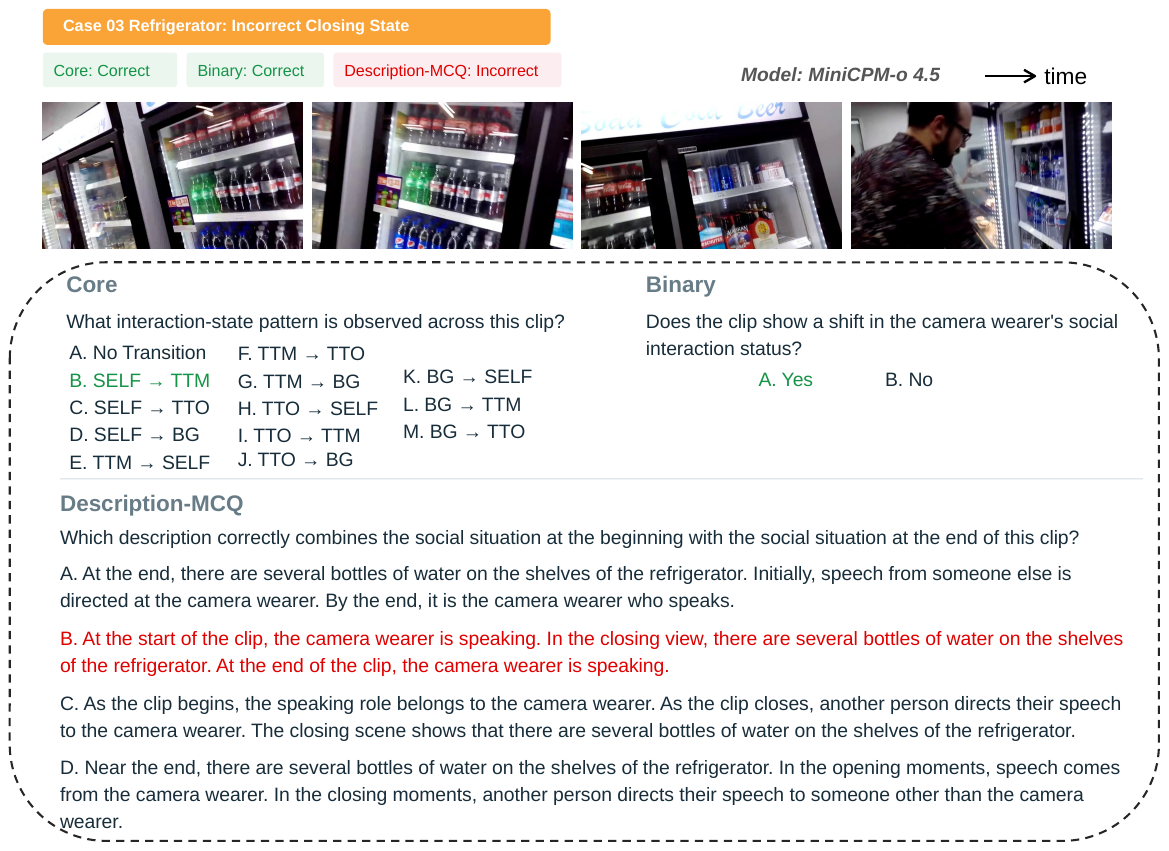}
    \caption{Interaction-state change example for MiniCPM-o 4.5.
    Core correctly identifies
    $\mathrm{SELF}\rightarrow\mathrm{TTM}$, and Binary
    correctly detects a change. The selected Description-MCQ
    option incorrectly states that the wearer speaks at both
    endpoints.
    Green and red indicate correct and incorrect
    model-selected answers, respectively.}
    \label{fig:supp_dynamics_case}
}

\paragraph{Interaction-state changes.}
Figure~\ref{fig:supp_dynamics_case} shows an example from
a refrigerator clip. The interaction changes from the wearer
speaking (SELF) to another person addressing the wearer (TTM).
The model correctly identifies
$\mathrm{SELF}\rightarrow\mathrm{TTM}$ in Core and detects
a change in Binary. Its selected Description-MCQ option,
however, states that the wearer is speaking at both the
beginning and the end. This description preserves the correct
initial state but misidentifies the closing state.
\newpage
\noindent
Thus, correct transition classification and change detection
do not ensure that the selected description consistently
represents the same event.

Together, these examples illustrate the complementary
diagnostic value of the three components. Core assesses
specific participant, state, or transition categories;
Binary tests targeted social propositions; and
Description-MCQ assesses whether multiple pieces of
information match within a single description.
Considering these predictions together reveals errors that
a correct answer in one component alone would leave unnoticed.

\FloatBarrier
\WideFigurePage{
\centering
    \includegraphics[clip,width=0.70\textwidth]{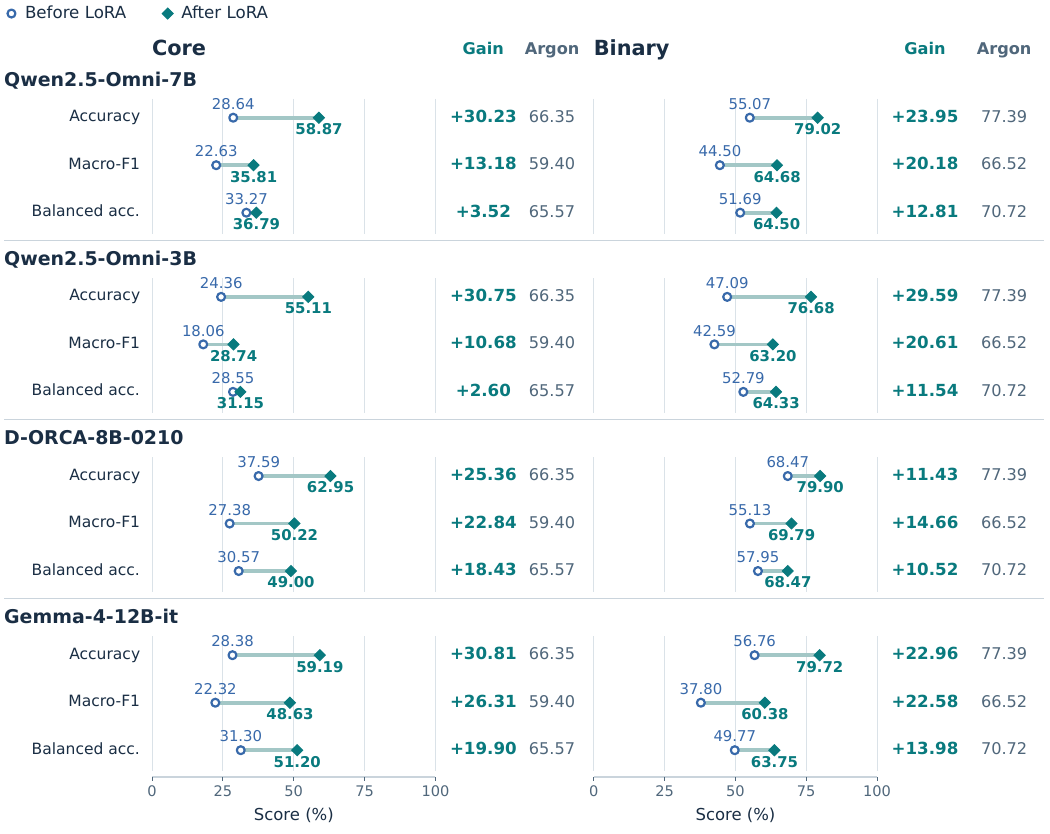}
    \caption{Core and Binary test scores before and after LoRA fine-tuning.
    Open circles denote base-model scores and filled diamonds denote
    fine-tuned scores. Gain is the change in percentage points (pp).
    The Argon columns give the direct-inference scores of Gemini-4-Argon.}
    \label{fig:supp_lora_metrics}
}

\subsection{Results Before and After LoRA Fine-Tuning}
\label{sec:supp_lora_results}

Figure~\ref{fig:supp_lora_metrics} compares Qwen2.5-Omni-7B,
Qwen2.5-Omni-3B, D-ORCA-8B-0210, and Gemma-4-12B-it on the same
test set before and after LoRA fine-tuning. We use the direct-inference
scores of Gemini-4-Argon as a reference. Following
Sec.~\ref{sec:supp_metrics}, we give each task equal weight when
averaging accuracy, balanced accuracy (BA), and macro-F1 over the
12 Core tasks and, separately, over the 9 Binary tasks.

\ResultParagraph{Similar accuracy gains, different gains in balanced metrics.}
All four models improve on all three metrics on both Core and Binary
after fine-tuning. On Core, Qwen2.5-Omni-3B and Gemma-4-12B-it gain
nearly the same amount in accuracy (30.75 and 30.81 percentage points,
respectively). Gemma gains 19.90 points in BA, which weights classes
equally within each task, compared with 2.60 points for Qwen-3B.
Gemma also gains more in macro-F1 (26.31 versus 10.68 points). Thus, similar gains in accuracy can
accompany substantially different gains in class-macro metrics.

\ResultParagraph{Fine-tuned D-ORCA versus Gemini-4-Argon.}
On Binary, fine-tuned D-ORCA exceeds Gemini-4-Argon by 2.51 points
in accuracy and 3.27 points in macro-F1, while its BA is 2.25 points
lower. On Core, D-ORCA approaches Argon in accuracy
(62.95\% vs.\ 66.35\%), but trails by 16.57 points in BA and
9.18 points in macro-F1. Thus, although fine-tuned D-ORCA approaches
Argon in task-macro Core accuracy, a substantial gap remains in
the class-macro metrics.
\par


\FloatBarrier

\end{document}